\documentclass{article} 
\usepackage{iclr2025_conference,times}

\usepackage{amsmath,amsfonts,bm}

\def\eqref#1{equation~\ref{#1}}

\def\1{\bm{1}}

\DeclareMathAlphabet{\mathsfit}{\encodingdefault}{\sfdefault}{m}{sl}
\SetMathAlphabet{\mathsfit}{bold}{\encodingdefault}{\sfdefault}{bx}{n}

\usepackage{hyperref}
\usepackage{url}

\usepackage{duckuments}
\usepackage{xspace}
\usepackage{xcolor}
\usepackage{graphicx}
\usepackage{blindtext}
\usepackage{algorithm}
\usepackage{algpseudocode}
\usepackage{algcompatible}
\usepackage{enumitem}
\usepackage{subcaption}
\usepackage{amsmath}
\usepackage{amsfonts}
\usepackage{mathtools}
\usepackage{booktabs}
\hypersetup{
  linkcolor=CubsRed,
  citecolor=CubsRed,
  urlcolor=CubsRed,
  colorlinks=true,
  filecolor=magenta, 
}
\usepackage{hyperref}       
\usepackage{bm}
\usepackage{wrapfig}
\usepackage{lipsum}
\usepackage{multirow}
\usepackage{cleveref}
\usepackage{tablefootnote}
\usepackage{tikz}
\usepackage{pifont}

\usepackage{listings}
\definecolor{Gray}{gray}{0.9}
\definecolor{Blue9}{rgb}{0.098,0.3,0.9}
\definecolor{Red7}{rgb}{0.941, 0.243, 0.243}
\definecolor{Green7}{RGB}{55, 178, 77}
\definecolor{CubsRed}{RGB}{198, 1, 31}
\definecolor{BrickRed}{rgb}{0.6,0,0}
\definecolor{RoyalBlue}{rgb}{0,0,0.8}
\definecolor{Tdgreen}{rgb}{0,0.4,0.7}
\definecolor{CYRed}{RGB}{228, 0, 43}
\definecolor{CYPurple}{RGB}{215, 153, 93}

\newcommand{\cmark}{\textcolor{Green7}{\ding{51}}}%
\newcommand{\xmark}{\textcolor{red}{\ding{55}}}%

\title{Subtask-Aware Visual Reward Learning from Segmented Demonstrations}
\newcommand{\metfull}{REward learning from Demonstration with Segmentations\xspace} 
\newcommand{\metabbr}{REDS\xspace}

\author{Changyeon Kim$^1${\quad}Minho Heo$^1{\quad}$Doohyun Lee$^1${\quad}\\
\textbf{Honglak Lee$^{2,3}${\quad}Jinwoo Shin$^1${\quad}Joseph J. Lim$^{1}$\thanks{Equal advising. \hfill \textbf{Project page:} {\scriptsize\url{https://changyeon.site/reds}}}{\quad}Kimin Lee$^{1}$$^\ast$}  \\
$^1$KAIST \,\,$^2$University of Michigan\,\,\,$^3$LG AI Research\,\,\, \\
}

\iclrfinalcopy 
\begin{document}

\maketitle

\begin{abstract}
    Reinforcement Learning (RL) agents have demonstrated their potential across various robotic tasks. However, they still heavily rely on human-engineered reward functions, requiring extensive trial-and-error and access to target behavior information, often unavailable in real-world settings. This paper introduces \metabbr: \textit{\metfull}, a novel reward learning framework that leverages action-free videos with minimal supervision. Specifically, \metabbr employs video demonstrations segmented into subtasks from diverse sources and treats these segments as ground-truth rewards. We train a dense reward function conditioned on video segments and their corresponding subtasks to ensure alignment with ground-truth reward signals by minimizing the Equivalent-Policy Invariant Comparison distance. Additionally, we employ contrastive learning objectives to align video representations with subtasks, ensuring precise subtask inference during online interactions. Our experiments show that \metabbr significantly outperforms baseline methods on complex robotic manipulation tasks in Meta-World and more challenging real-world tasks, such as furniture assembly in FurnitureBench, with minimal human intervention. Moreover, \metabbr facilitates generalization to unseen tasks and robot embodiments, highlighting its potential for scalable deployment in diverse environments.

\end{abstract}

\section{Introduction}
\label{sec:introduction}
    Reinforcement Learning (RL) has demonstrated significant potential for training autonomous agents in various real-world robotic tasks, provided that appropriate reward functions are available~\citep{levine2016end, gu2017deep, andrychowicz2020learning, smith2022walk, handa2023dextreme}. However, reward engineering typically requires substantial trial-and-error~\citep{booth2023perils, knox2023reward} and extensive task knowledge, often necessitating specialized instrumentation (\textit{e.g.}, motion trackers~\citep{peng2020learning} or tactile sensors~\citep{yuan2023robot}) or detailed information about target objects~\citep{james2020rlbench, robosuite2020, yu2020meta, mu2021maniskill, gu2023maniskill, sferrazza2024humanoidbench}, which are difficult to obtain in real-world settings. Learning reward functions from action-free videos has emerged as a promising alternative, as it avoids the need for detailed action annotations or precise target behavior information, and video data can be easily collected from online sources~\citep{soomro2012ucf101, kay2017kinetics, Damen2018EPICKITCHENS}. Approaches in this domain include learning discriminators between video demonstrations and policy rollouts~\citep{chen2021learning, r2r}, training temporally aligned visual representations from large-scale video datasets~\citep{sermanet2018time, zakka2022xirl, kumar2023graph, ma2023vip, ma2023liv} to estimate reward based on distance to a goal image, and using video prediction models to generate reward signals~\citep{VIPER, diffusion_reward}.

    Despite this progress, existing methods often struggle with long-horizon, complex robotic tasks that involve multiple subtasks. These approaches typically fail to provide context-aware reward signals, relying only on a few consecutive frames or the final goal image without considering subsequent subtasks. For example, in One Leg task (see Figure~\ref{fig:example_one_leg}) from FurnitureBench~\citep{furniturebench}, prior methods often overemphasize the reward for picking up the leg while neglecting crucial steps such as inserting the leg into a hole and tightening it. Recent work~\citep{drs} proposes a discriminator-based approach that treats complex tasks as a sequence of subtasks. However, it assumes that the environment provides explicit subtask identification, which often demands significant human intervention in real-world scenarios. Moreover, discriminator-based methods are known to be prone to mode collapse~\citep{wang2017robust, TRAIL} (please refer to Figure~\ref{fig:baseline_qual_fb} for empirical evidence over prior work). Consequently, designing an effective visual reward function for real-world, long-horizon tasks remains an open problem.

\begin{figure*}[t]
    \centering
    \includegraphics[width=\textwidth]{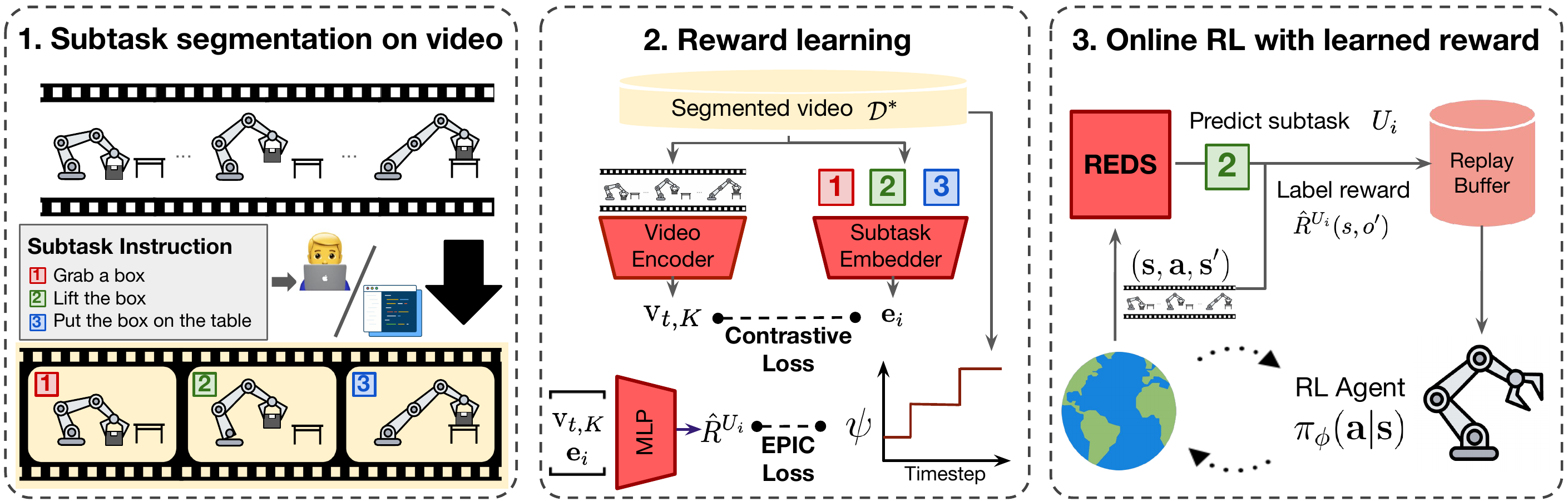}
    \caption{
    Illustration of \metabbr. Our main idea is to leverage expert demonstrations annotated with the ongoing subtask as the source of implicit reward signals (left). We train a reward model conditioned on video segments and corresponding subtasks with 1) contrastive loss to attract the video segments and corresponding subtask embeddings and 2) EPIC~\citep{EPIC} loss to generate reward equivalent to subtask segmentations (middle). In online RL, \metabbr infers the ongoing subtask using only video segments at each timestep and computes the reward with that (right).
    }
    \vspace{-0.175in}
    \label{fig:overview}
\end{figure*}

    \paragraph{Our approach} To address the aforementioned limitations, we propose a novel reward learning framework, \metabbr: \textit{\metfull}, which infers subtask information from video segments and generates corresponding reward signals for each subtask. The key idea is to employ minimal supervision to produce appropriate reward signals for intermediate subtask completion. Specifically, \metabbr utilizes expert demonstrations, where subtasks are annotated at each timestep by various sources (e.g., human annotators, code snippets, vision-language models; see the left figure of Figure~\ref{fig:overview}). These annotations serve as ground-truth rewards. For training, we introduce a new objective function minimizing the Equivalent-Policy Invariant Comparison (EPIC)~\citep{EPIC} between the learned reward function and the ground-truth rewards, guaranteeing a theoretical upper bound on regret relative to the ground-truth reward function. Additionally, to correctly infer the ongoing subtask in online interactions, we adopt a contrastive learning objective to align video representations with task embeddings. In terms of architecture, our reward model is designed to capture temporal dependencies in video segments using transformers~\citep{vaswani2017attention}, leading to enhanced reward signal quality.

    We find that \metabbr can generate appropriate reward signals to solve complex tasks by recognizing subtask structures, enabling the agent to efficiently explore and solve tasks through online interactions, using only expert demonstrations and subtask segmentations. 
    Our experiments show that RL agents trained with \metabbr achieve substantially improved sample efficiency compared to baseline methods on various robotic manipulation tasks from Meta-World~\citep{yu2020meta}. Additionally, we show that \metabbr can effectively train agents to perform long-horizon, complex furniture assembly tasks from FurnitureBench~\citep{furniturebench} using real-world online RL. Moreover, \metabbr facilitates RL training in unseen environments involving new tasks and embodiments, which would otherwise require significant effort in prior reward-shaping methods.
    
    \vspace{-0.06in}
    
    \paragraph{Contributions} We highlight the key contributions of our paper below:
    \vspace{-0.06in}
    \begin{itemize}[leftmargin=4mm]
        \item We present a novel visual reward learning framework \metabbr: \textit{\metfull}, which can produce suitable reward signals aware of subtasks in long-horizon complex robotic manipulation tasks.
        \item We show that \metabbr significantly outperforms baselines in training RL agents for robotic manipulation tasks in Meta-world, and even surpasses dense reward functions in some tasks.
        \item We demonstrate that \metabbr can train real-world RL agents to perform long-horizon complex furniture assembly tasks from FurnitureBench.
        \item We demonstrate that our approach shows strong generalization across various unseen tasks, embodiments, and visual variations.
    \end{itemize}

\section{Related Work}
\label{sec:related_work}
    
    \paragraph{Reward learning from videos} 
    Learning from observations without expert actions has been a promising research area because it does not require extensive instrumentation and allows for the easy collection of vast amounts of video from online sources. Notably, several studies have proposed methods for learning rewards directly from videos and using the signal to train RL agents. Previous work has been focused on learning a reward function by aligning video representations in temporal order \citep{sermanet2018time, zakka2022xirl, kumar2023graph} while others train a reward function for expressing the progress of the agent towards the goal \citep{Hartikainen2020Dynamical, lee2021generalizable, r2r}. Most recent work~\citep{VIPER} inspired by the success of video generative models~\citep{yan2021videogpt, ho2022video} utilizes the likelihood of pre-trained video prediction models as a reward. To effectively utilize video for long-horizon tasks, we propose a new reward model conditioned both on video segments and corresponding subtasks trained with subtask segmentations. 
    
    \paragraph{Inverse reinforcement learning} Designing an informative reward function remains a long-standing challenge for training RL agents. To achieve this, Inverse Reinforcement Learning (IRL) ~\citep{ng2000algorithms, abbeel2004apprenticeship, ziebart2008maximum} aims to estimate the underlying reward function from expert demonstrations. Adversarial imitation learning (AIL) approaches~\citep{GAIL, fu2018learning, oril, TRAIL, drs} address this by training a discriminator network to discriminate transitions from expert data or policy rollouts and using the output from the discriminator as a reward for training agents with RL. The most similar work to ours is DrS~\citep{drs}, which also utilizes subtask information of the multi-stage task. While DrS assumes that the information on ongoing subtasks can be obtained from the environment during online interaction, our method has no such assumption, so it can be applied in more general cases when the segmenting of the subtask is hard in automatic ways (e.g.,~\cite {furniturebench}).
    \vspace{-0.05in}

\section{Preliminaries}
\label{sec:preliminaries}
    \paragraph{Problem formulation} We formulate a visual control task as a Markovian Decision Process (MDP)~\citep{sutton2018reinforcement}. As a single image observation is not sufficient for fully describing the underlying state of the task, we use the set of consecutive past observations to approximate the current state following common practice~\citep{mnih2015human, yarats2021image, yarats2022mastering}. Taking this into account, we define MDP as a tuple $\mathcal{M} = (\mathcal{S}, \mathcal{A}, p, R, \rho_0, \gamma)$. $\mathcal{S}$ is a state space consisting of a stack of $K$ consecutive images,
    
    $\mathcal{A}$ is an action space, $R$ is the sparse reward function which outputs 1 when the agent makes success; otherwise, 0, $p(s'|s, a)$ is the transition function, $\rho_0$ is the initial state distribution, and $\gamma$ is the discount factor. The policy $\pi: \mathcal{S} \rightarrow \Delta(\mathcal{A})$ is trained to maximize the expected sum of discounted rewards $\mathbb{E}_{\rho_0, \pi, p}\left[\sum_{t=0}^{\infty} \gamma^{t}R(s_t, a_t)\right]$. Our goal is to find a dense reward function $\hat{R}(s')$ only conditioned on visual observations, from which we can get the optimal policy $\pi^{*}$ for $\mathcal{M}$. 
    
    \vspace{-0.05in} 
    
    \paragraph{EPIC} Equivalent-Policy Invariant Comparison (EPIC)~\citep{EPIC} is a pseudometric for quantifying differences between different reward functions, which is designed to ensure the invariance on the equivalent set of reward functions inducing the same set of optimal policies. To this end, EPIC first canonicalizes potential shaping of the reward function $R$ with some arbitrary distribution $\mathcal{D}_{\mathcal{S}} \in \Delta(\mathcal{S})$ over states $\mathcal{S}$, which is to be invariant to potential shaping as below\footnote{We only consider the action-independent reward functions, and omit the prime notation on $s'$ for the simplicity of notation.}:
    \begin{equation}
        C_{\mathcal{D}_{\mathcal{S}}}(R)(s) = R(s) + \mathbb{E}_{S \sim \mathcal{D}_{\mathcal{S}}}[ \gamma R(S) - R({S}) - \gamma R(S) ] = R(s) - \mathbb{E}_{S \sim \mathcal{D}_{\mathcal{S}}}\left[ R(S) \right],
    \end{equation}
    where $S$ denotes a set of batches independently sampled from the arbitrary distribution $\mathcal{D}_\mathcal{S}$, and $\gamma$ is the discount factor. EPIC is then defined by the Pearson distance between canonically shaped rewards in a scale-invariant manner:
    \begin{equation}
        D^{\text{EPIC}}_{\mathcal{D}_{\mathcal{C}}, \mathcal{D}_{\mathcal{S}}}(R_A, R_B) = \mathbb{E}_{s \sim \mathcal{D}_{\mathcal{C}}}\left[ D_{\rho}(C_{\mathcal{D}_{\mathcal{S}}}(R_A)(s), C_{\mathcal{D}_{\mathcal{S}}}(R_B)(s))\right],
    \end{equation}
    where $s$ is from the coverage distribution $\mathcal{D}_{\mathcal{C}}$, $D_{\rho}(X,Y) = \sqrt{\frac{1 - \rho(X, Y)}{2}}$ is the Pearson distance between two random variables $X$ and $Y$, and $\rho(X, Y)$ is the Pearson correlation between $X$ and $Y$. Please refer to \citet{EPIC} for more details.
    
\section{Method}
\label{sec:method}
    \vspace{-0.06in}

    This section presents \metabbr: \textit{\metfull}, a visual reward learning framework designed for long-horizon tasks involving multiple subtasks. To generate proper reward signals for solving intermediate subtasks, we utilize segmentations identifying ongoing subtasks in demonstrations. In Section~\ref{sec:segmentation}, we explain our intuition and formal definitions behind subtask segmentation. Section~\ref{sec:arch} outlines the reward model architecture, and Section~\ref{sec:reward_modeling} describes our training objective. Finally, in Section~\ref{sec:reward_modeling}, we elaborate on the details of training and inference of REDS. For an overview, see Figure~\ref{fig:overview}.
    
    \vspace{-0.06in}
    \subsection{Subtask Segmentation}
    \vspace{-0.06in}
    \label{sec:segmentation}
    The sparse reward function $R$ provides feedback only on the overall success or failure of a task, which is insufficient for guiding the agent through intermediate states. To address this, drawing inspiration from previous work on long-horizon robotic manipulation tasks~\citep{di2022learning, mandlekar2023mimicgen, furniturebench, drs}, we decompose a task into $m$ object-centric subtasks, denoted as $\mathcal{U} = \{U_{1}, ..., U_{m}\}$. Each subtask $U_i$ represents a distinct step in the task sequence and is based on the coordinate frame of a single target object. \footnote{For instance, Door Open can be divided into (i) reaching the door handle (which involves motion relative to the door handle.)} and (ii) pulling the door to the goal position (which involves motion relative to the green sphere-shaped goal). This approach is intuitive because humans naturally perceive tasks as sequences of discrete object interactions, and this assumption can be generally applied to different manipulation skills (e.g., pick-and-place, inserting) with diverse objects. Additionally, we provide text instructions $\mathcal{X} = \{\bm{x}_{i}\}_{i=1}^{m}$ that describe how to solve each subtask, which helps guide the agent more effectively. 
 
    To obtain subtask segmentations, we map each observation $o_t$ at timestep $t$ in the trajectory $\tau = (o_0, ..., o_T)$ to its corresponding subtask using a segmentation function $\psi: \mathcal{O} \rightarrow \mathcal{U}$. Specifically, $\psi$ outputs the index of the ongoing subtask based on the observation at each timestep, with the output value increasing as the number of completed subtasks increases (refer to the graph of $\psi$ in the center of Figure~\ref{fig:overview}). The function $\psi$ can be derived from various sources such as code snippets based on domain knowledge~\citep{james2022q, james2022coarse, mees2022calvin}, guidance from human teachers~\citep{furniturebench}, or vision-language models~\citep{zhang2024universal, koukisa}. In our experiments, we use the predefined codes in Meta-world and human annotators in FurnitureBench to collect subtask segmentations. Note that these segmentations are only used during training; our framework is designed to automatically infer subtasks during online interactions without external annotations.

    \vspace{-0.06in}

    \subsection{Architecture}
    \vspace{-0.06in}
    \label{sec:arch}
        
        As mentioned in Section~\ref{sec:introduction}, previous reward learning methods generate rewards only by a single frame or consequent frames, not taking into account the order of subtasks. To resolve the issue, we propose a new reward predictor $\hat{R}^{U} = \hat{R}(s; U)$ conditioned on each subtask. To efficiently process visual observations, we first encode each image into low-dimensional representations using a pre-trained visual encoder $E_v$. To capture temporal dependencies, these representations are processed through a causal transformer~\citep{vaswani2017attention}. We add positional embeddings for each image in the sequence $s_t$ and pass them through the transformer network, producing the output representation $\bm{v}_{t,K} = \{\bm{v}_{t-K-1}, ..., \bm{v}_{t-1}, \bm{v}_{t}\}$ such that $t$-th output depends on input up to $t$. To embed subtask ${U}_i$, we encode $\bm{x}_{i}$ with pre-trained text encoder $E_t$ and project it to a shallow MLP to earn $\bm{e}_{i}$.
        This design allows \metabbr to generate rewards for unseen tasks when $\mathcal{U}$ and $\mathcal{X}$ are provided. (see Section~\ref{exp:generalization} for supporting experiments). Finally, we concatenate a sequence of video representations $\bm{v}_{t,K}$ and subtask embedding $\bm{e}_{i}$ to $[\bm{v}_{t-K-1}, ... \bm{v}_{t}, \bm{e}_{i}]$ and project it to another shallow MLP $f$ to obtain $\hat{R}_{\theta}(s_{t};U_{i}) = f(\bm{v}_{t,K}, \bm{e}_{i})$. 
        \vspace{-0.06in}
        
    \subsection{Reward Modeling}
    \vspace{-0.06in}
         
        \label{sec:reward_modeling} 
        \paragraph{Reward equivariance with subtask segmentation} Our key insight is that the subtask segmentation function $\psi$ can be thought of as the ground-truth reward function, providing implicit signals for solving intermediate tasks. To ensure our reward function induces the same set of optimal policies as $\psi$, we train to minimize EPIC~\citep{EPIC} distance between our reward model $\hat{R}^{U}_{\theta}$ parameterized by $\theta$ and $\psi$ for all subtasks:
        \begin{equation}
            \label{eq:epic}
            \mathcal{L}_{\text{EPIC}}(\theta) = \frac{1}{k} \sum_{i=1}^{k}D^{\text{EPIC}}_{\mathcal{D}_{\mathcal{C}}, \mathcal{D}_{\mathcal{S}}}(\hat{R}^{U_{i}}_{\theta}, \psi).
        \end{equation}
        
        \paragraph{Progressive reward signal} However, minimizing EPIC with $\psi$ alone can lead to overfitting and the inability to provide progressive signals within each subtask. To mitigate this issue, we propose an additional regularization term to enforce progressive reward signals. Inspired by previous work~\citep{lee2021generalizable, Hartikainen2020Dynamical, wu2021learning}, we view the reward function as a progress indicator for each subtask, and we regularize the reward function output to be higher in later states of expert demonstration as follows:
        \begin{align}
            \begin{split}
            \mathcal{L}_{\text{reg}}(\theta) = \max\left(0, \epsilon - (\hat{R}_{\theta}(s_{t+j}; \psi(o_{t+j})) - \hat{R}_{\theta}(s_{t}; \psi(o_{t})))\right),
            \end{split}
            \label{eq:progressive}
        \end{align}

        where $j$ is randomly chosen from a fixed set of values, and $\epsilon$ is a hyperparameter. Note that we apply this objective only for the expert demonstrations and not suboptimal demonstrations collected in iterative processes (please refer to Section~\ref{sec:train_infer}).

        \vspace{-0.06in}
        \paragraph{Aligning video representation with subtask embeddings} As the reward model lacks information about the ongoing subtasks in online interactions, it must infer the agent's current subtask. To achieve this, we train the video representation to be closely aligned with the corresponding subtask embedding by adopting a contrastive learning objective to make the model select the appropriate subtask embedding only by the video segment. 
        \begin{equation}
            \label{eq:cont}
            \mathcal{L}_{\text{cont}}(\theta) = -\log\frac{\mathrm{sim}(\bm{v}_{t},\bm{e}_{\psi(o_t)})}{\sum_{i \in \{1, ..., k\}}\mathrm{sim}(\bm{v}_{t}, \bm{e}_{i})},
        \end{equation}

        where $\mathrm{sim}$ represents a cosine similarity. 
        
        In summary, all components parameterized by $\theta$ are jointly optimized to minimize our total training objective:
        \begin{equation}
            \label{eq:total_obj}
            \mathcal{L}(\theta) =  \mathcal{L}_{\text{EPIC}}(\theta) + \mathcal{L}_{\text{reg}}(\theta) + \mathcal{L}_{\text{cont}}(\theta).
        \end{equation}
        
        \vspace{-0.06in}
        
    \subsection{Training and Inference}
        \label{sec:train_infer}
        \vspace{-0.06in}
        \paragraph{Inference} To compute the reward at timestep $t$, REDS first infers the agent's current subtask from a sequence of recent visual observations. Specifically, REDS selects the subtask index $\bar{i}$ by choosing the subtask embedding $\bm{e}_{\bar{i}}$ that has the highest cosine similarity with the final output of the causal transformer, denoted as $\bm{v}_t$. The final reward is then computed using video embedding and text embedding of the inferred subtask as $\hat{R}_{\theta}(s_{t}; U_{\bar{i}}) = f(\bm{v}_t, \bm{e}_{\bar{i}})$. Please refer to Appendix~\ref{appendix:impl} for more details.
        \vspace{-0.06in}

        \paragraph{Training} We outline the training procedure for REDS. First, we collect subtask segmentations from expert demonstrations, creating a dataset $\mathcal{D}^{0}$, and use it to train the initial reward model, $M^0$. However, reward models trained solely on expert data are susceptible to reward misspecification~\citep{pan2022the}. To address this, we iteratively collect suboptimal demonstrations and fine-tune the reward model using expert and suboptimal data. Unlike expert demonstrations, suboptimal demonstrations cover a broader range of states and more diverse observations, making manual segmentation labor-intensive and error-prone. To reduce the burden on human annotators, we develop an automatic subtask inference procedure, avoiding the need for manual segmentation.

        Before the iterative process, we compute similarity scores for all states in the expert demonstrations using the initial reward model $M^0$. For each subtask $U_i$, we calculate a threshold $T_{U_i}$ based on the similarity scores between the expert states and the corresponding instructions, ensuring $T_{U_i}$ represents the minimum similarity required for successful subtask completion. In each iteration $i \in \{1, ..., n\}$, we proceed as follows:
        \begin{itemize}[leftmargin=4mm]
            \item Step 1 (Suboptimal data collection): We train an RL agent using the reward model $M^{i}$ and collect suboptimal demonstrations $\mathcal{D}^{i}_{\text{replay}}$ from the agent's replay buffer.
            \item Step 2 (Subtask inference for suboptimal data): For each timestep in the suboptimal trajectory, we infer the subtask index $\hat{i}$ using the same procedure as in inference and compute $\mathrm{sim}(\bm{v}_t, \bm{e}_{\hat{i}})$. If the similarity falls below the threshold $T_{U_{i}}$ at any timestep, we mark the subtask as failed and assign the remaining timesteps to that subtask.
            \item Step 3 (Fine-tuning): We fine-tune the reward model $M^{i-1}$ using the combined dataset $\mathcal{D}^{i} = \mathcal{D}^{i} \cup \mathcal{D}^{i}_{\text{replay}}$ to obtain $M^{i}$.
        \end{itemize}
        \vspace{-0.06in}
        We use the final reward model $M^{n}$ for downstream RL training.

\begin{figure}
    \centering
    \begin{subfigure}[b]{0.17\textwidth}
        \centering
        \includegraphics[width=\textwidth]{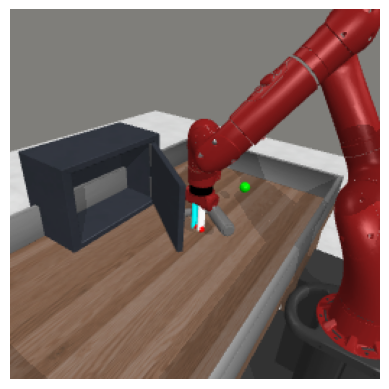}
        \caption[]{Door Open}    
        \label{fig:example_door_open}
    \end{subfigure}
    \begin{subfigure}[b]{0.17\textwidth}   
        \centering 
        \includegraphics[width=\textwidth]{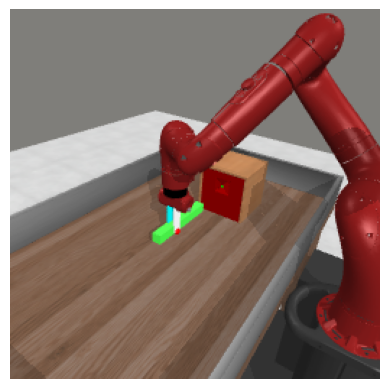}
        \caption[]{Peg Insert Side}    
        \label{fig:example_peg_insert_side}
    \end{subfigure}
    \begin{subfigure}[b]{0.17\textwidth}  
        \centering 
        \includegraphics[width=\textwidth]{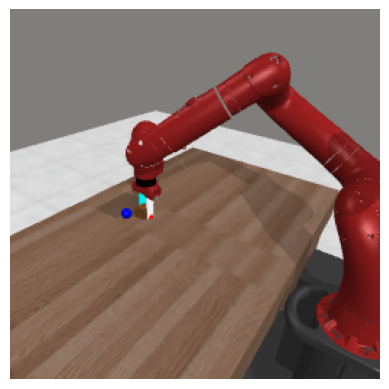}
        \caption[]{Sweep Into}    
        \label{fig:example_sweep_into}
    \end{subfigure}
    \begin{subfigure}[b]{0.34\textwidth}   
        \centering 
        \includegraphics[width=\textwidth]{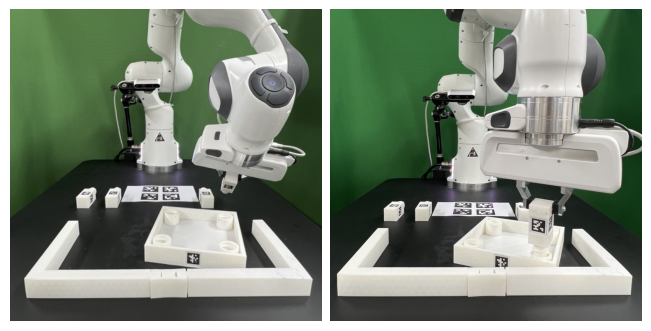}
        \caption[]{One Leg}    
        \label{fig:example_one_leg}
    \end{subfigure}
 
    \caption{Examples of visual observations used in our experiments. We consider a variety of robotic manipulation tasks from Meta-world~\citep{yu2020meta} and FurnitureBench~\citep{furniturebench}.} 
    \label{fig:mean and std of nets}
    \vspace{-0.2in}
\end{figure}
\section{Experiments}
\vspace{-0.06in}
\label{sec:exp}
We design our experiment to evaluate the effectiveness of \metabbr on providing useful reward signals in training various RL algorithms~\citep{hafner2023mastering, kostrikov2022offline}. We conduct extensive experiments in robotic manipulation tasks from Meta-world~\citep{yu2020meta} (see Section~\ref{exp:mw}) in simulation and robotic furniture assembly tasks from FurnitureBench~\citep{furniturebench} (see Section~\ref{exp:fb}) in the real-world. We also conduct in-depth analyses to validate the effectiveness of each component and how our reward function aligns with subtask segmentations (see Section~\ref{exp:abl}).
\vspace{-0.06in}
    \paragraph{Implementation and training details} We used the open-source pre-trained CLIP~\citep{CLIP} with ViT-B/16 architecture to encode images and subtask instructions for all experiments. We adopt a GPT~\citep{radford2018improving} architecture with 3 layers and 8 heads for the causal transformer. To canonicalize our reward functions, we use the same $\mathcal{D}$ for both coverage distribution $\mathcal{D}_{\mathcal{C}}$ and potential shaping distribution $\mathcal{D}_\mathcal{S}$, and we estimate the expectation over state distributions using a sample-based average over 8 additional samples from $\mathcal{D}$ per sample. All models are trained with AdamW~\citep{loshchilov2018decoupled} optimizer with a learning rate of $1 \times 10^{-4}$ and a mini-batch size of 32. To ensure to visual distractions, we apply color jittering and random shift~\citep{yarats2021image} to visual observations in training \metabbr. Please refer to Appendix~\ref{appendix:impl} for more details.
    \vspace{-0.06in}
    
    \paragraph{Baselines} We consider the following baselines: (1) human-engineered reward functions provided in the benchmark, (2) ORIL~\citep{oril}, an adversarial imitation learning (AIL) method trained only with offline demonstrations,  (3) Rank2Reward (R2R)~\citep{r2r}, an AIL method which trains a discriminator weighted with temporal ranking of video frames to reflect task progress, (4) VIPER~\citep{VIPER}, a reward model utilizing likelihood from a pre-trained video prediction model as a reward signal, and (5) DrS~\citep{drs}, an AIL method that assumes subtask information from the environment and trains a separate discriminator for each subtask. We provide additional details on baselines in Appendix~\ref{appendix:baseline}.   
    \vspace{-0.06in}
    
    \subsection{Meta-world Experiments}
    \label{exp:mw}
        \paragraph{Setup} We first evaluate our method on 8 different visual robotic manipulation tasks from Meta-world~\citep{yu2020meta}. As a backbone algorithm, we use DreamerV3~\citep{hafner2023mastering}, a state-of-the-art visual model-based RL algorithm that learns from latent imaginary rollouts. For collecting subtask segmentations, we utilize a scripted teacher in simulation environments for scalability. Specifically, we use the predefined indicator for subtasks provided in the benchmark for all subtask segmentations (see Appendix~\ref{appendix:task} for the list of subtasks and corresponding text instructions for each task). We do not use these indicators when training/evaluating RL agents. For training \metabbr, we first collect subtask segmentations from 50 expert demonstrations for initial training and train DreamerV3 agents for 100K environment steps with the initial reward model to collect suboptimal trajectories, which is used for fine-tuning. 
        In evaluation, we measure the success rate averaged over 10 episodes in every 20K steps. Please refer to Appendix~\ref{appendix:impl} for more details.

\begin{figure}[t]
    \centering
    \begin{subfigure}[b]{\linewidth}
        \centering
        \includegraphics[width=0.8\linewidth]{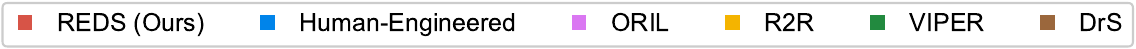}
    \end{subfigure}
    \begin{subfigure}[b]{0.245\textwidth}
        \centering
        \includegraphics[width=\textwidth]{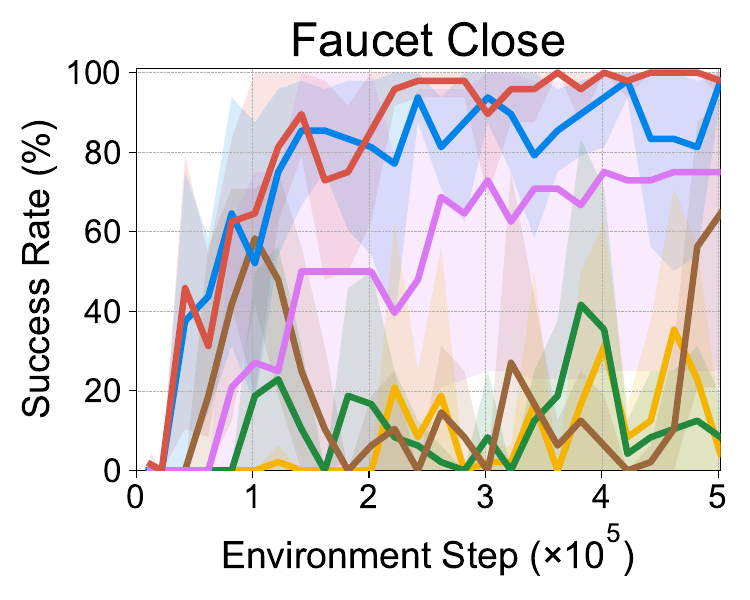}
        \label{fig:mw_faucet}
        \vspace{-0.175in}
    \end{subfigure}
    \begin{subfigure}[b]{0.245\textwidth}  
        \centering 
        \includegraphics[width=\textwidth]{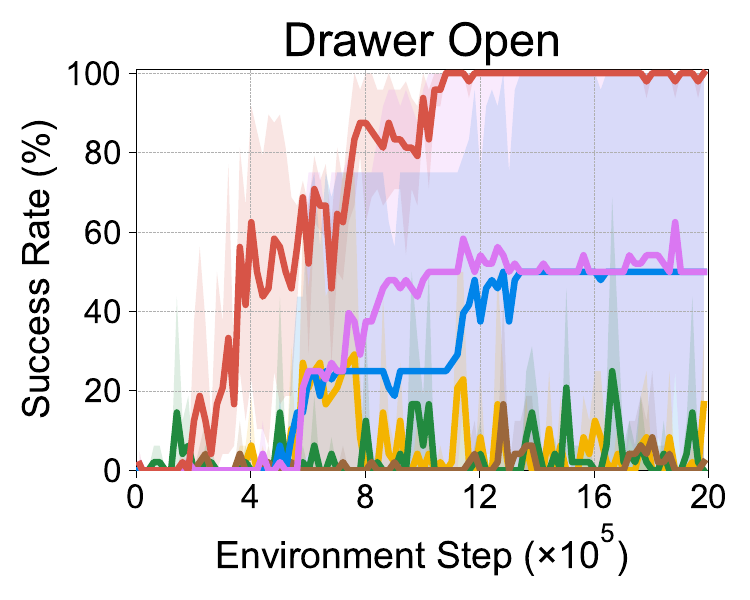}
        \label{fig:mw_drawer}
        \vspace{-0.175in}
    \end{subfigure}
    \begin{subfigure}[b]{0.245\textwidth}   
        \centering 
        \includegraphics[width=\textwidth]{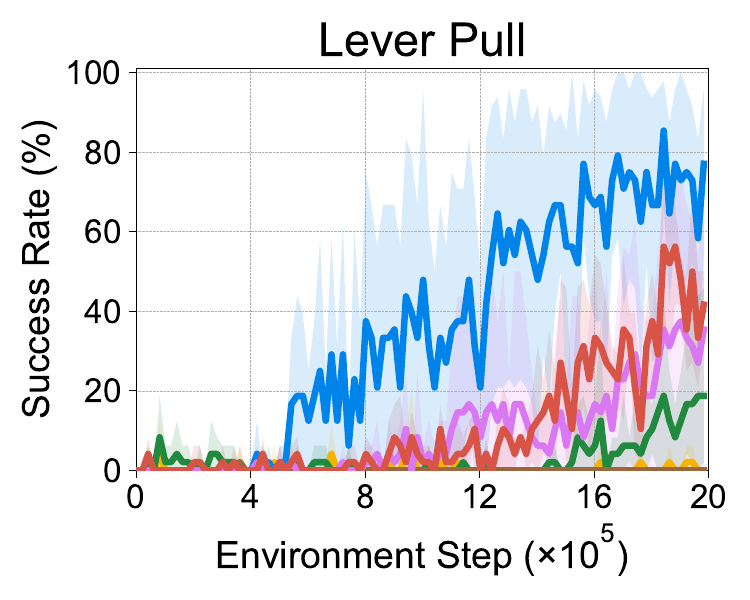}
        \label{fig:mw_lever}
        \vspace{-0.175in}
    \end{subfigure}
    \begin{subfigure}[b]{0.245\textwidth}   
        \centering 
        \includegraphics[width=\textwidth]{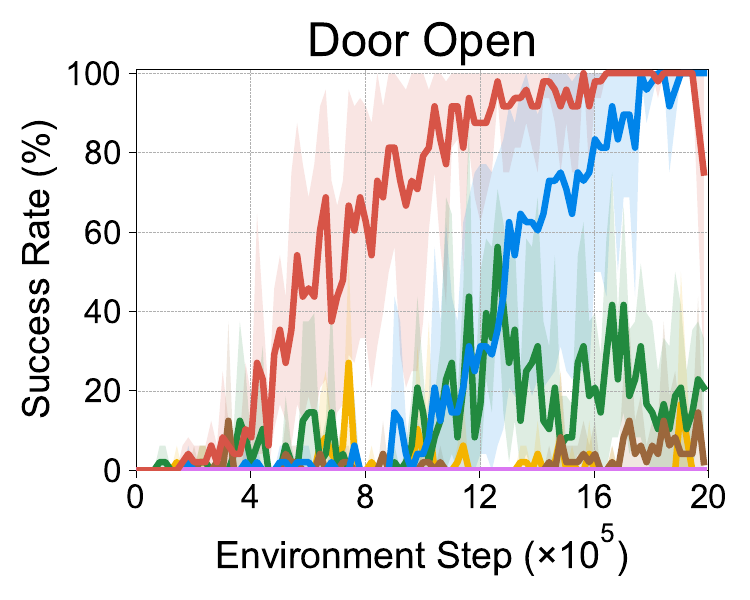}
        \label{fig:mw_door}
        \vspace{-0.175in}
    \end{subfigure}
    \begin{subfigure}[b]{0.245\textwidth}   
        \centering 
        \includegraphics[width=\textwidth]{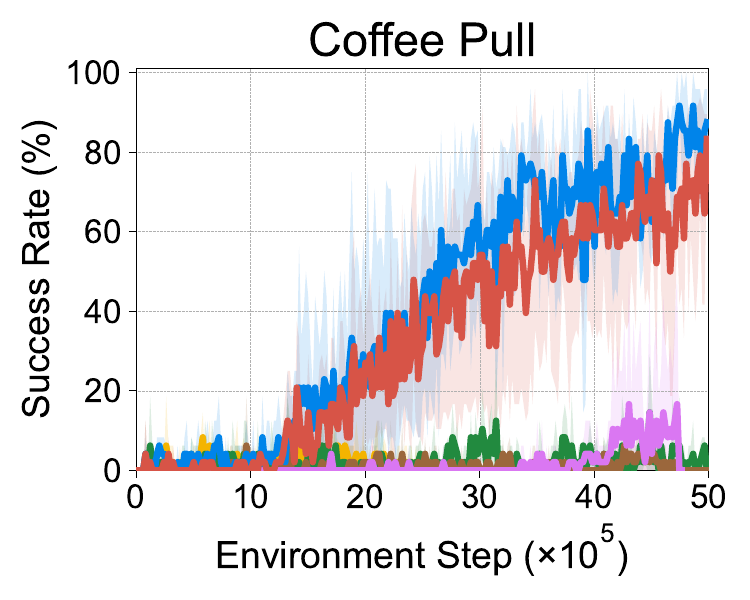}
        \label{fig:mw_coffee_pull}
        \vspace{-0.175in}
    \end{subfigure}
    \begin{subfigure}[b]{0.245\textwidth}   
        \centering 
        \includegraphics[width=\textwidth]{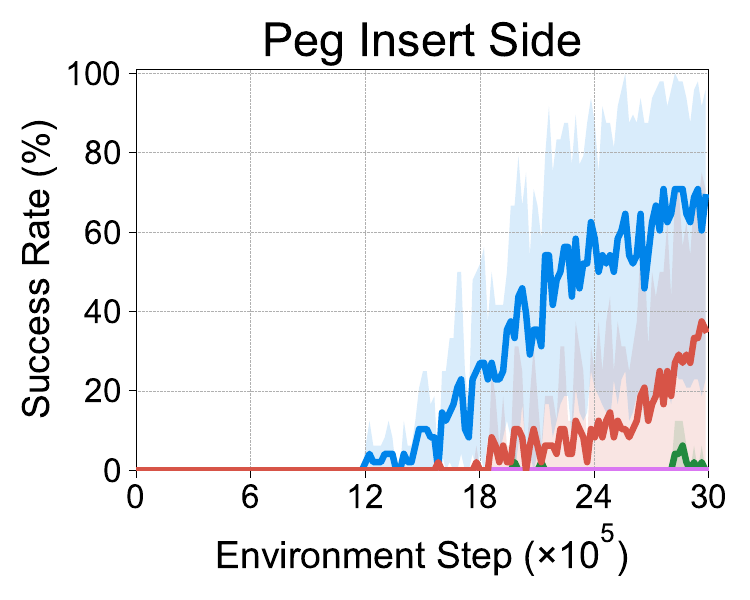}
        \label{fig:mw_peg_insert}
        \vspace{-0.175in}
    \end{subfigure}
    \begin{subfigure}[b]{0.245\textwidth}
        \centering
        \includegraphics[width=\textwidth]{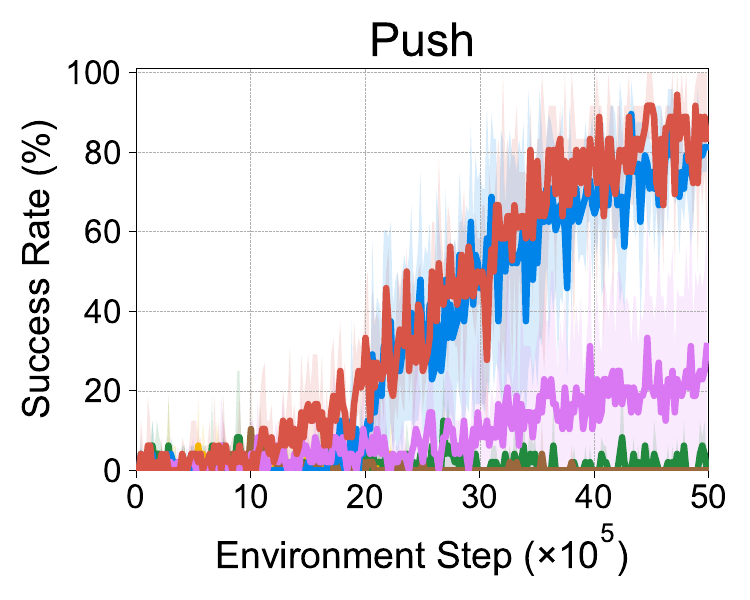}
        \label{fig:mw_push}
        \vspace{-0.175in}
    \end{subfigure}
    \begin{subfigure}[b]{0.245\textwidth}  
        \centering 
        \includegraphics[width=\textwidth]{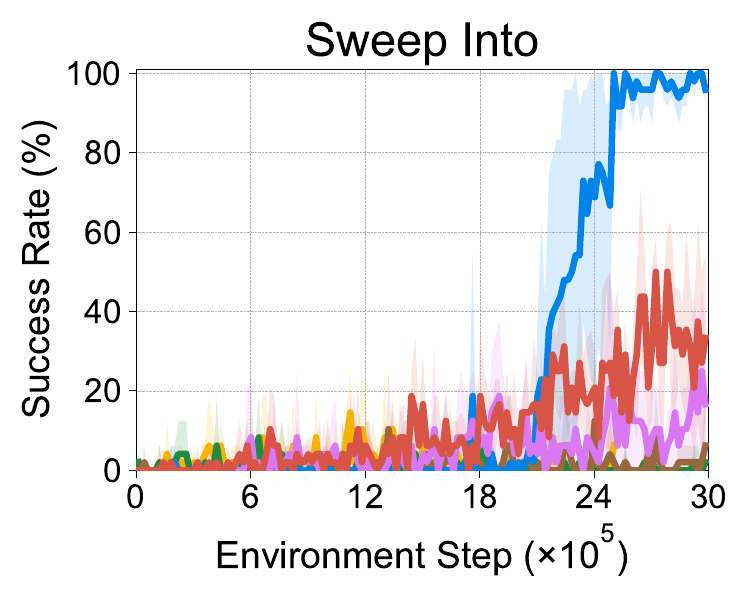}
        \label{fig:mw_sweep_into}
        \vspace{-0.175in}
    \end{subfigure}

    \caption{
    Learning curves of DreamerV3~\citep{hafner2023mastering} agents trained with different reward functions for solving eight robotic manipulation tasks from Meta-world~\citep{yu2020meta}, measured by success rate (\%). The solid line and shaded regions represent the mean and stratified bootstrap interval across 4 runs. \textcolor{cyan}{}
    }
    \label{fig:metaworld}
   \vspace{-0.2in}
\end{figure}

        \vspace{-0.1in}
        \paragraph{Results} Figure~\ref{fig:metaworld} shows that \metabbr consistently improves the sample-efficiency of DreamerV3 agents by outperforming all baselines. While baselines exhibit non-zero success rates in simple tasks like Faucet Close, their performance significantly deteriorates in more complex tasks, such as Peg Insert Side. On the other hand, our method maintains non-zero success rates across all tasks and even surpasses human-engineered reward functions in some tasks (e.g., Drawer Open, Push, Coffee Pull) without requiring task-specific reward engineering. These results show that \metabbr effectively generates appropriate rewards for solving intermediate tasks by leveraging subtask-segmented demonstrations. A key advantage of \metabbr is that it relies solely on visual observations for generating rewards during online interaction, whereas DrS and human-engineered rewards require additional information from the environment, such as the position and reachability of target objects. This result underscores \metabbr's potential for application in environments where reward engineering is challenging or additional sensory information is unavailable. 

        \vspace{-0.1in}
        
    \subsection{FurnitureBench Experiments}
    \label{exp:fb}
\begin{wrapfigure}{r}{0.52\textwidth}
    \vspace{-0.15in}
    \captionof{table}{
        Online fine-tuning results of IQL agents in One Leg from FurnitureBench. We report the initial performance after offline RL (left) and the performance after 150 episodes of online RL (right).
    }\label{tbl:furniturebench}
    \centering\small\resizebox{0.52\textwidth}{!}{
        \begin{tabular}{@{}lcc@{}}
        \toprule
        \multirow{2}{*}{Method} & \multirow{2}{*}{\# Expert Demos} & Completed Subtasks                   \\
                                & & (Offline $\rightarrow$ Online)       \\ \midrule
        Sparse (Offline)~\citep{furniturebench} & 500 & 1.8\\
        VIPER                   & 300 & 1.10 $\rightarrow$ 1.25            \\
        DrS             & 300 & 1.05 $\rightarrow$ 1.10            \\
        \textbf{REDS (Ours)}             & \textbf{300} & \textbf{1.10 $\rightarrow$ 2.45} \\

        \bottomrule
        \end{tabular}
    }
    \vspace{-0.1in}
\end{wrapfigure}

        \vspace{-0.06in}
        \paragraph{Setup} We further evaluate our method on real-world furniture assembly tasks from FurnitureBench~\citep{furniturebench}, specifically focusing on One Leg assembly.
        This task involves a sequence of complex subtasks such as picking up, inserting, and screwing (see Figure~\ref{fig:example_one_leg}). For training REDS, we use 300 expert demonstrations with subtask segmentations provided by FurnitureBench, along with an additional 200 rollouts from IQL~\citep{kostrikov2022offline} policy trained with expert demonstrations in a single training iteration. To prevent misleading reward signals stemming from visual occlusions, we utilize visual observations from the front camera and wrist cameras in training \metabbr.
        For downstream RL, we first train offline RL agents using 300 expert demonstrations labeled with each reward model, followed by online fine-tuning to assess improvements. For baselines, we compare against VIPER and DrS. We emphasize that our method enables fully autonomous training in online RL sessions, in contrast to DrS, which relies on a subtask indicator provided by humans. In our DrS experiments, subtasks were manually identified by a human. We measure the average number of completed subtasks over 20 rollouts for evaluation. We provide more details in Appendix~\ref{appendix:impl}.
        \vspace{-0.05in}
        \paragraph{Results} As shown in Table~\ref{tbl:furniturebench}, \metabbr achieves significant performance improvements through online fine-tuning, whereas the improvements from baselines are marginal. These results indicate that our method produces informative signals for solving a sequence of subtasks, while baselines either fail to provide context-aware signals or dense rewards for better exploration (see Appendix~\ref{appendix:qualitative} for qualitative examples). Moreover, we note that our method outperforms the IQL trained with 500 expert demonstrations, achieving a score of 2.45 compared to 1.8  reported by FurnitureBench, despite using only 300 expert demonstrations.
        Considering that \metabbr does not require additional human interventions beyond resetting the environment, these results highlight the potential to extend our approach to a wider range of real-world robotics tasks.
        
        \vspace{-0.1in}

    \subsection{Alignment with Ground-truth Rewards}
    \label{sec:alignment}
        \vspace{-0.06in}
\begin{wrapfigure}{r}{0.55\textwidth}
    \vspace{-0.15in}
    \captionof{table}{
        EPIC~\citep{EPIC} distance (lower is better) between learned reward functions and hand-engineered reward functions (Meta-world) / subtask segmentations (FurnitureBench) in unseen data. 
    }\label{tbl:epic}
    \vspace{-0.05in}
    \centering\resizebox{0.55\textwidth}{!}{
        \begin{tabular}{@{}lcccc@{}}
        \toprule
            Task               & VIPER  & R2R & ORIL   & \metabbr (Ours)            \\ 
        \midrule
        Meta-world Door Open       & 0.5934 & 0.5649 & 0.7071 & \textbf{0.4913}          \\
        Meta-world Push            & 0.6144 & 0.6838 & 0.7073 & \textbf{0.5381} \\
        Meta-world Peg Insert Side & 0.5974 & 0.5806 & 0.6989 & \textbf{0.4674}         \\
        Meta-world Sweep Into      & 0.6248 & 0.6413 & 0.7001 & \textbf{0.4673}                \\
        FurnitureBench One Leg     & 0.7035 & 0.6001 & 0.7014 & \textbf{0.0713}            \\ 
        \bottomrule
        \end{tabular}
    }
    \vspace{-0.1in}
\end{wrapfigure}
\begin{figure} [t] \centering
    \begin{subfigure}[b]{.48\textwidth}
         \centering
         \includegraphics[width=\textwidth]{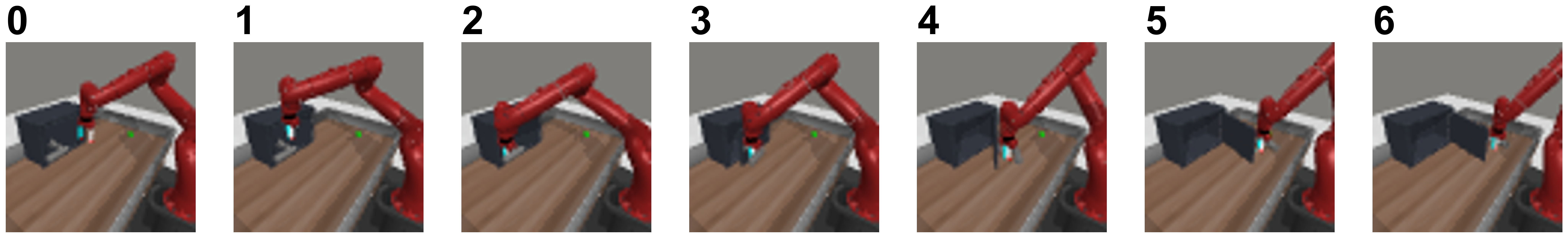}
         \includegraphics[width=\textwidth]{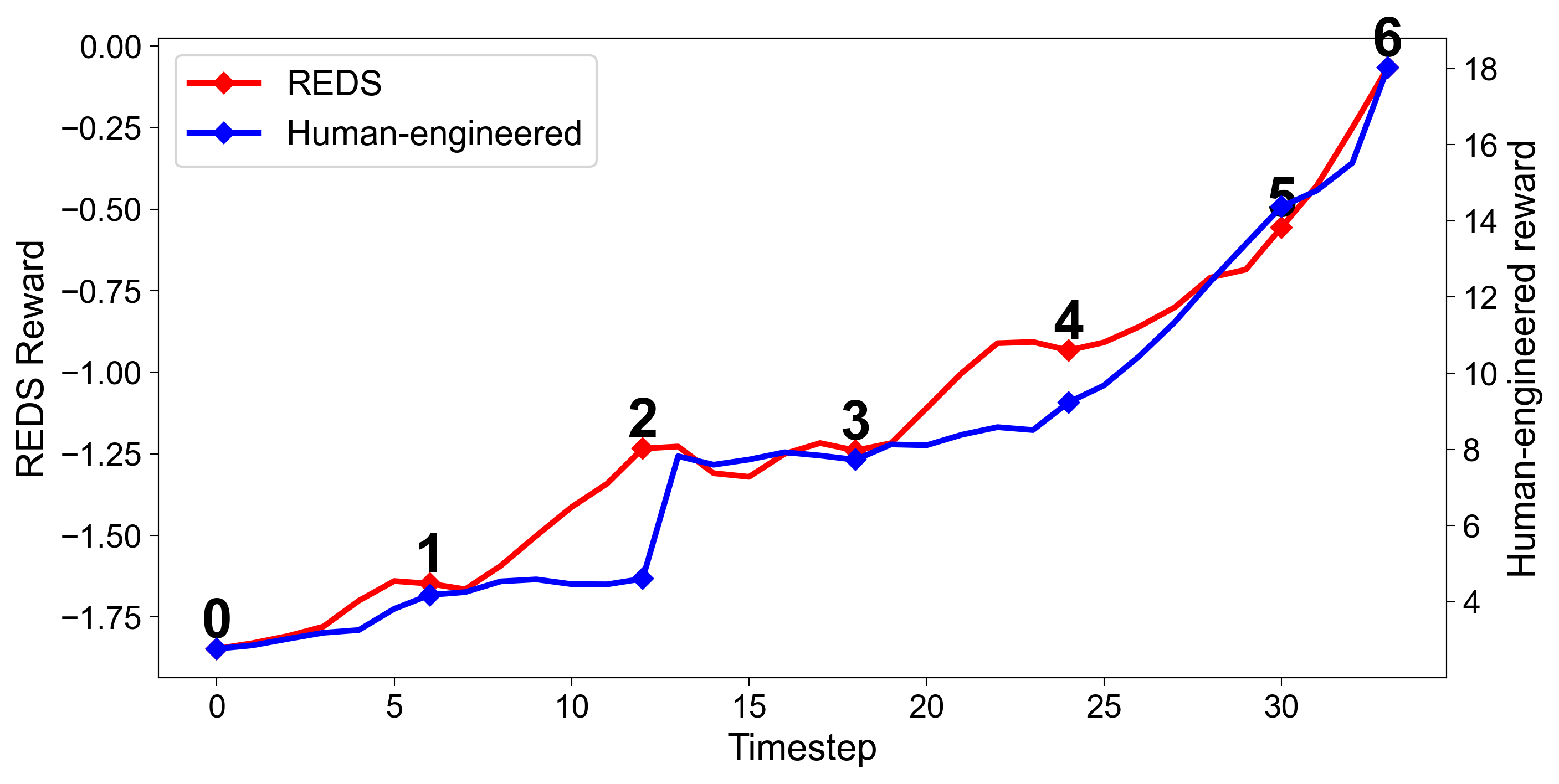}
         \label{fig:qual_main_mw_door_open}
         \vspace{-0.1in}
         \caption{Door Open}
    \end{subfigure}
    \begin{subfigure}[b]{.48\textwidth}
         \centering
         \includegraphics[width=\textwidth]{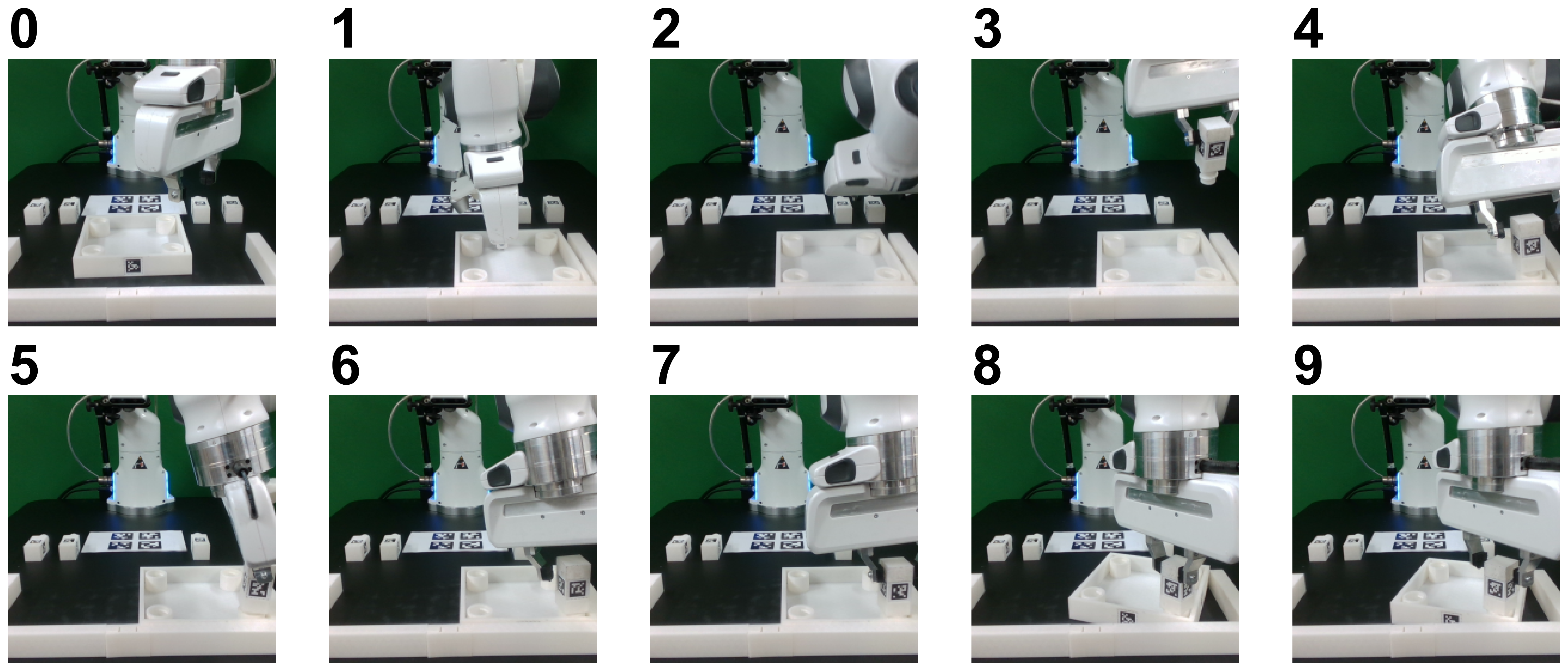}
         \includegraphics[width=\textwidth]{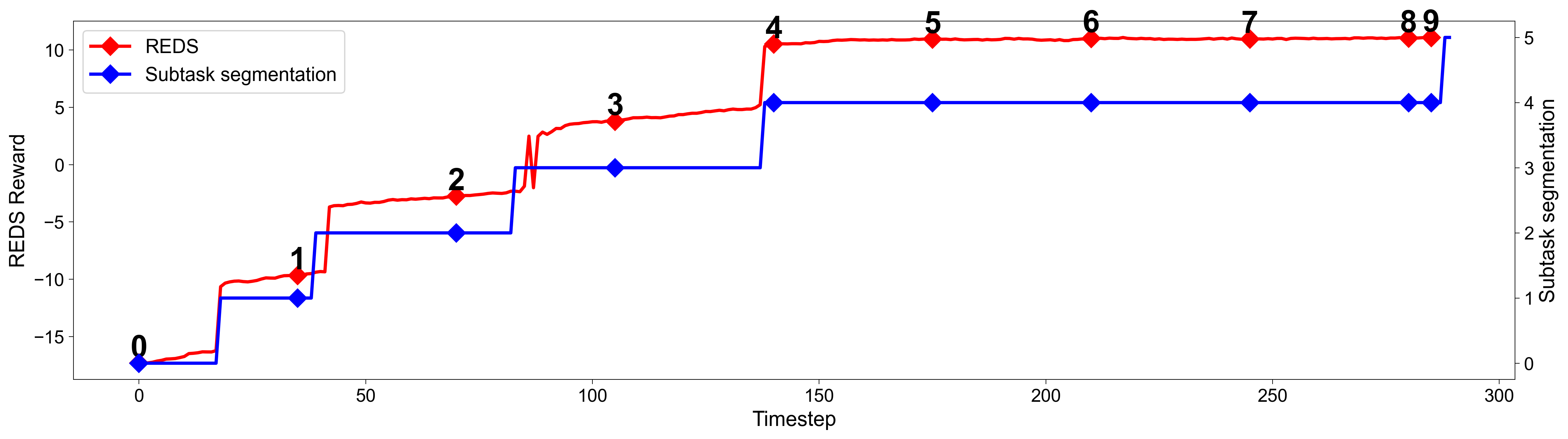}
         \vspace{-0.1in}
         \label{fig:qual_main_fb_one_leg}
         \caption{One Leg}
    \end{subfigure}

    \vspace{-0.05in}
    \caption{
    Qualitative results of \metabbr in Door Open in Meta-world~\citep{yu2020meta} and One Leg from FurnitureBench~\citep{furniturebench}. We observe that \metabbr produces suitable reward signals aligned with ground-truth reward functions by predicting ongoing subtasks effectively and providing progressive reward signals.
    }
    \label{fig:qual_main}
    \vspace{-0.1in}
\end{figure}

        \paragraph{EPIC measurement} To quantitatively validate the alignment of our method with ground-truth reward functions, we measure the EPIC distance with a set of unseen demonstrations during training. Specifically, we use rollouts from the reference policy trained with expert demonstrations for state distribution. In Table~\ref{tbl:epic}, we observe that \metabbr exhibits significantly lower EPIC distance than baselines across all tasks. Particularly, the difference between \metabbr and baselines is more pronounced in complex tasks like One Leg. This result consistently supports the empirical findings from previous sections.
        \vspace{-0.06in}
        \paragraph{Qualitative analysis} We provide the graph of computed rewards from REDS in Figure~\ref{fig:qual_main}. We observe that \metabbr can induce suitable reward signals aligned with ground-truth reward functions. For example, \metabbr provides subtask-aware signals in transition states (e.g., between 2 and 3, and between 4 and 5) and generates progressive reward signals throughout each subtask. Please refer to Appendix~\ref{appendix:qualitative} for the extensive comparison between \metabbr and baselines. 
        \vspace{-0.1in}

\begin{figure}[t]
    \centering
    \begin{subfigure}[b]{0.33\textwidth}
        \centering
        \includegraphics[width=\textwidth]{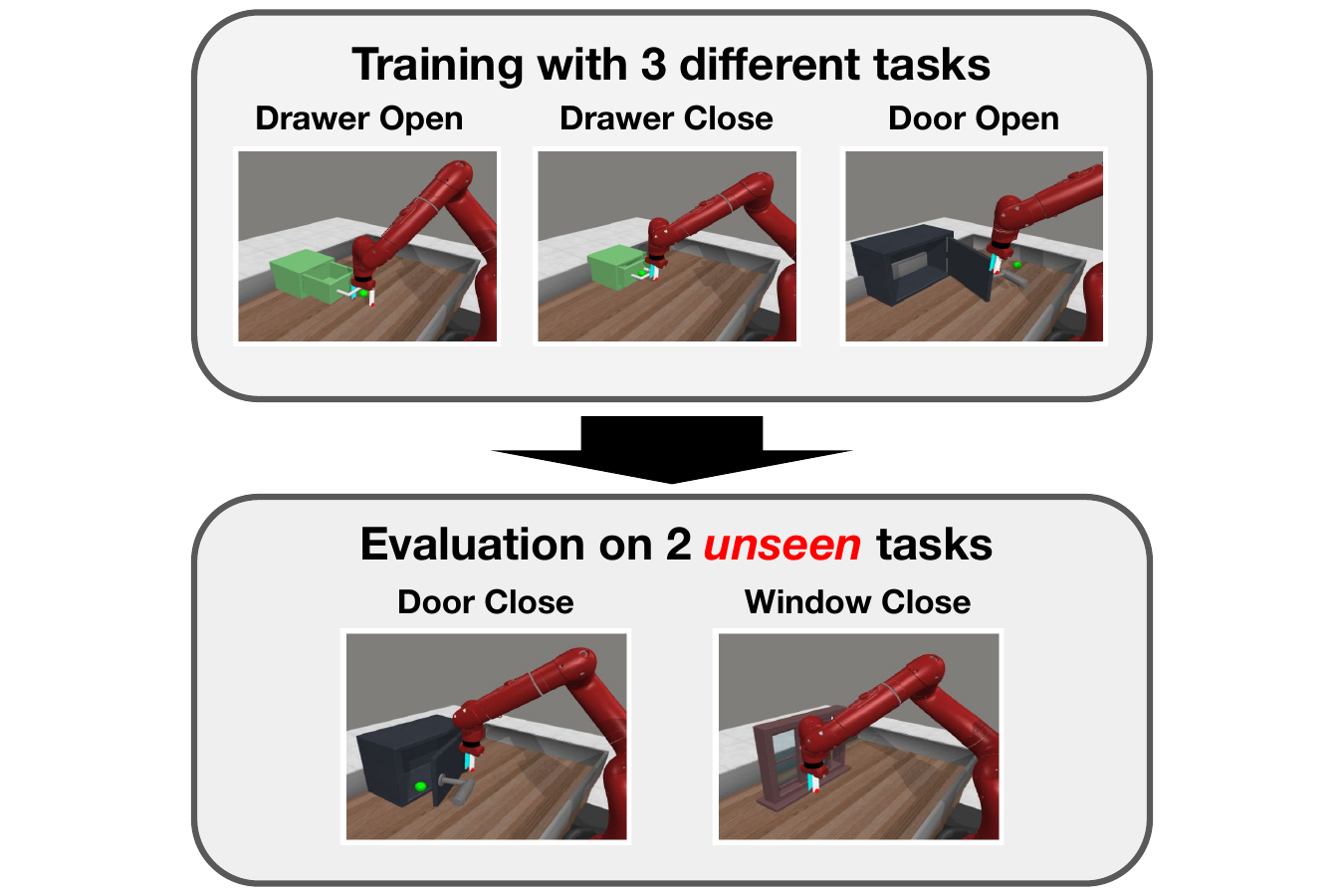}
        \label{fig:gen_concept}
        \vspace{-0.15in}
    \end{subfigure}
    \begin{subfigure}[b]{0.28\textwidth}
        \centering
        \includegraphics[width=\textwidth]{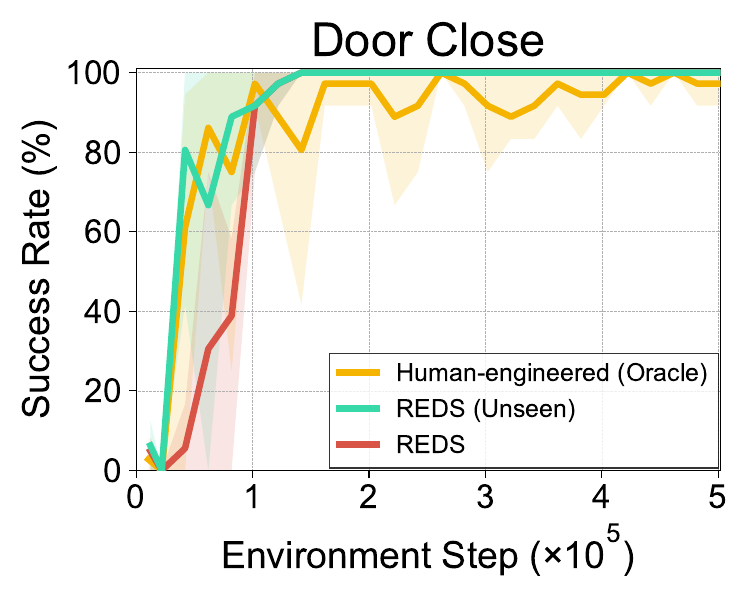}
        \vspace{-0.15in}
        \label{fig:gen_door_close}
    \end{subfigure}
    \begin{subfigure}[b]{0.28\textwidth}  
        \centering 
        \includegraphics[width=\textwidth]{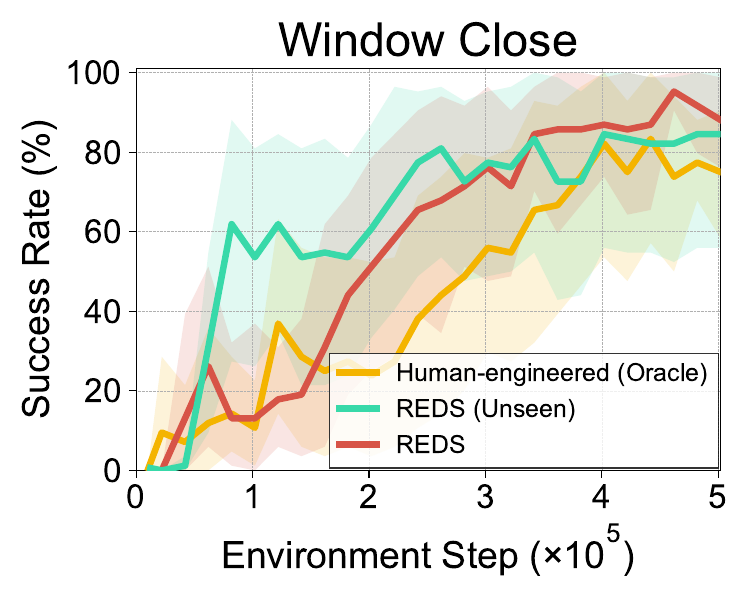}
        \vspace{-0.15in}
        \label{fig:gen_window_close}
    \end{subfigure}

    \caption{
        We train \metabbr with 3 different tasks from Meta-world~\citep{yu2020meta} and use this model to train RL agents in 2 unseen tasks (left). We present learning curves on Door Close (center) and Window Close (right), as measured by success rate (\%). The solid line and shaded regions represent the mean and stratified bootstrap interval across 4 runs.
    }
    \label{fig:generalization}
    \vspace{-0.15in}
\end{figure}

    \subsection{Generalization Capabilities}
    \label{sec:generalization}
        \vspace{-0.06in}
        \paragraph{Transfer to unseen tasks} As mentioned in Section~\ref{sec:arch}, our model can be applied as a reward function in unseen tasks. To validate this, we conduct additional experiments by training \metabbr with segmentation data from 3 tasks (Door Open, Drawer Open/Close) and using the reward model to train RL agents in two unseen tasks. In Door Close, we aim to validate that \metabbr can provide informative signals for a new task involving a previously seen object and behaviors. In Window Close, we aim to determine whether \metabbr can provide suitable reward signals for familiar behaviors (closing) with an unseen object (window). In evaluation, we change the text instruction following the target object (as shown in Table~\ref{appendix:task}), and we do not fine-tune the reward model. Figure~\ref{fig:generalization} shows that \metabbr provides effective reward signals on unseen tasks and achieves comparable or even better RL performance than \metabbr trained on the target task. This result demonstrates that \metabbr can be applied to RL training in unseen tasks that share properties with training tasks.  
        \label{exp:generalization}
        \vspace{-0.1in}

        \begin{figure}[t]
    \centering
    \begin{subfigure}[b]{0.15\textwidth}
        \centering
        \includegraphics[width=\textwidth]{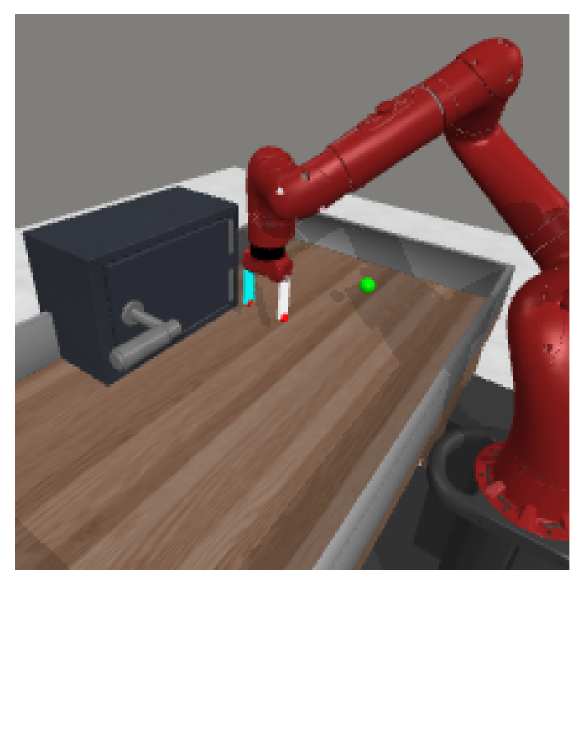}
        \vspace{-0.25in}
        \caption[]{Original}    
        \label{fig:example_original}
    \end{subfigure}
    \begin{subfigure}[b]{0.5\textwidth}   
        \centering 
        \includegraphics[width=\textwidth]{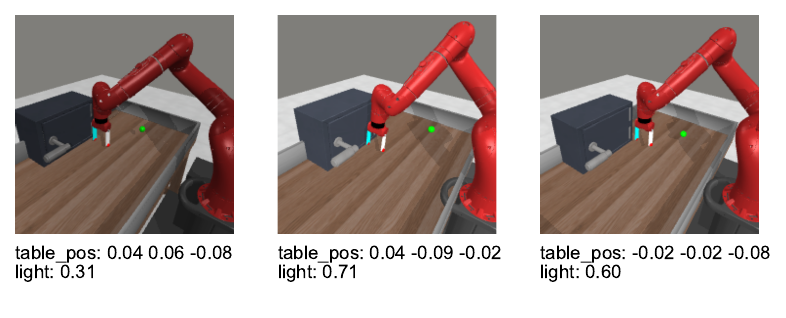}
        \vspace{-0.25in}
        \caption[]{Examples of visual distractions}    
        \label{fig:visual_distraction_examples}
    \end{subfigure}
    \begin{subfigure}[b]{0.27\textwidth}  
        \centering 
        \includegraphics[width=\textwidth]{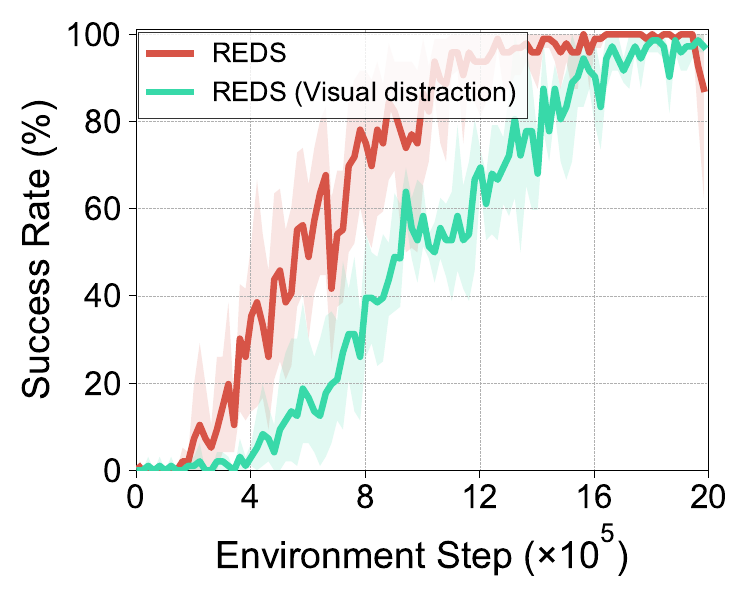}
        \vspace{-0.25in}
        \caption[]{Learing curve}
        \label{fig:visual_distraction_graph}
    \end{subfigure}
    \vspace{-0.05in}

    \caption{
    We provide visual observations from (a) the original environment and (b) unseen environments with visual distractions used in our experiments in Section~\ref{sec:generalization}
.    }
    \label{fig:visual_distraction}
    \vspace{-0.2in}
\end{figure}
        \paragraph{Robustness to visual distractions} To prove the robust performance of \metabbr against visual distractions, we train RL agents with our reward model in new Meta-world environments incorporating visual distractions, such as varying light and table positions following \cite{xie2023decomposing} (see Figure~\ref{fig:visual_distraction_examples}). Note that the reward model was trained using demonstrations only from the original environment. As Figure~\ref{fig:visual_distraction_graph} shows, \metabbr can generate robust reward signals despite visual distractions and train RL agents to solve the task effectively.
        \vspace{-0.06in}

\begin{wrapfigure}{r}{0.3\textwidth}
    \vspace{-0.15in}
    \captionof{figure}{
        Learning curve for DreamerV3 agents in environments of the Sawyer Arm. 
    }\label{fig:embodiment_graph}
    \vspace{-0.15in}
    \begin{center}
        \includegraphics[width=0.3\textwidth]{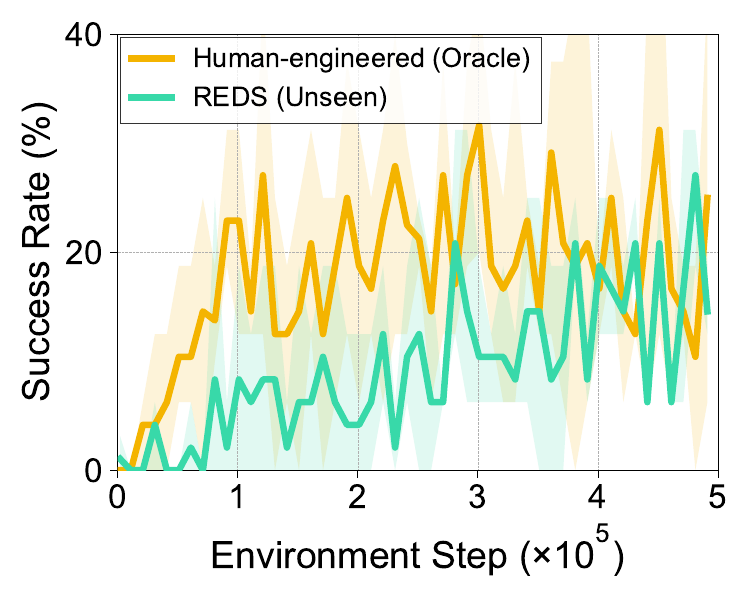}
    \end{center}
    \vspace{-0.26in}
\end{wrapfigure}
\begin{figure} [ht] \centering
    \begin{subfigure}[b]{.45\textwidth}
         \centering
         \includegraphics[width=\textwidth]{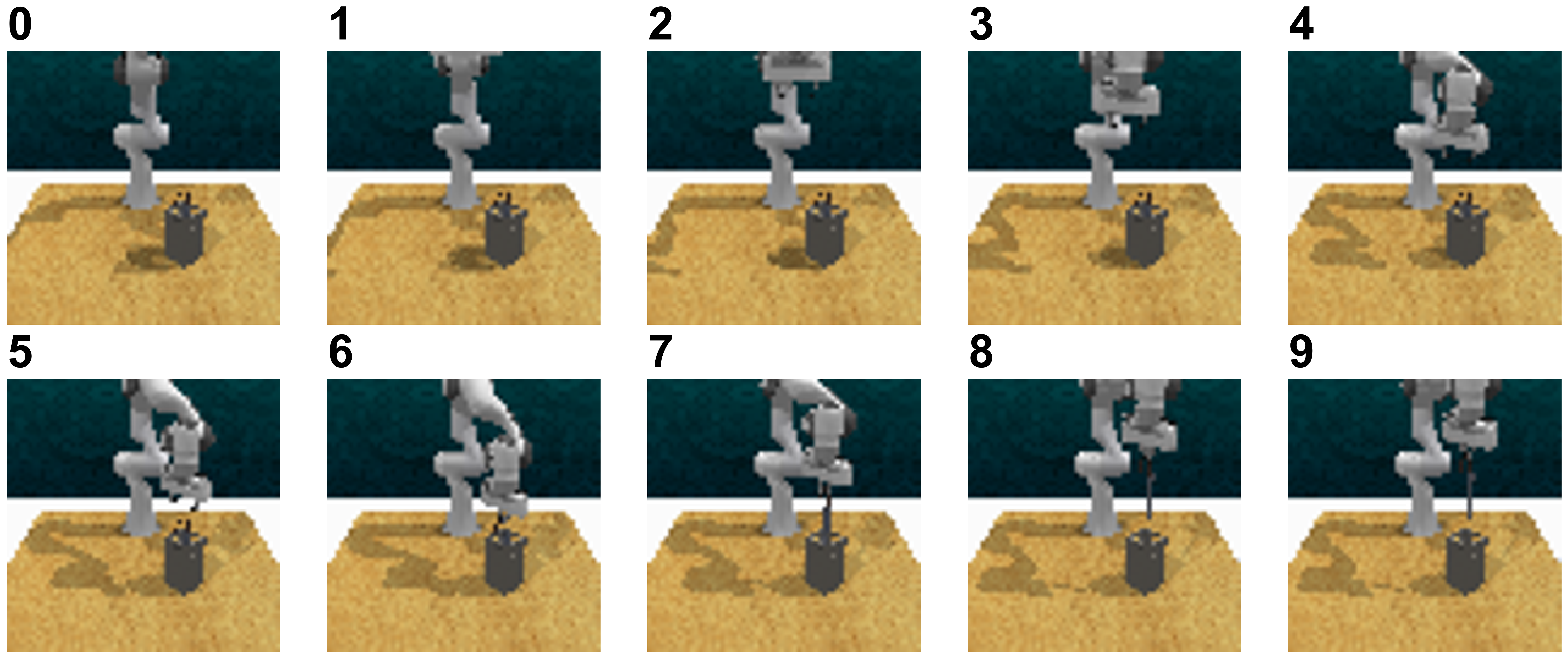}
         \includegraphics[width=\textwidth]{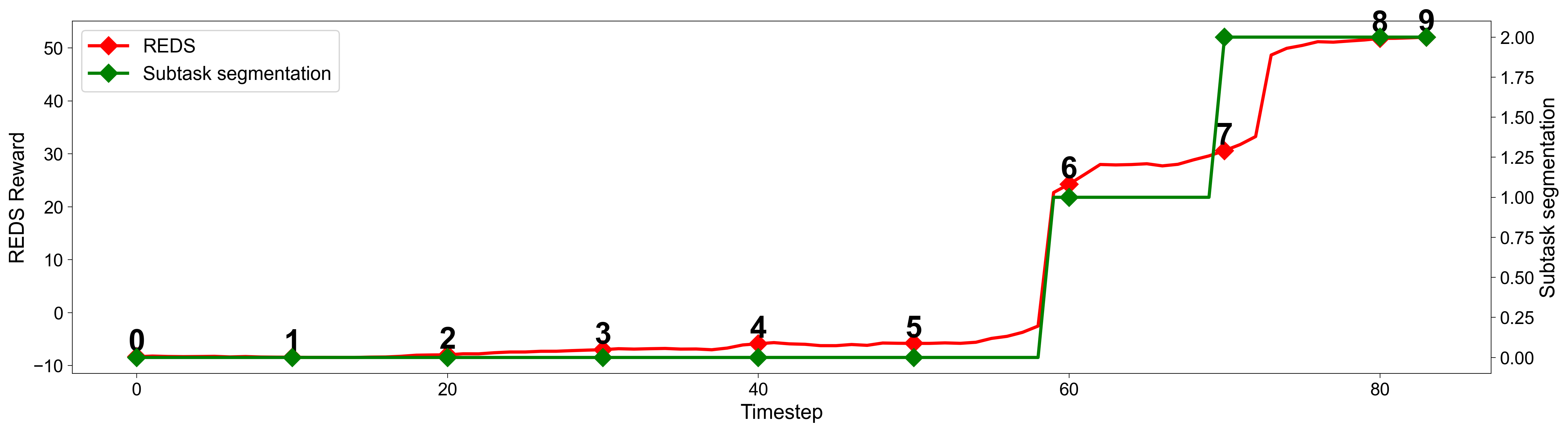}
         \label{fig:reds_rlbench_panda}
         \vspace{-0.15in}
         \caption{(Train) Panda}
    \end{subfigure}
    \begin{subfigure}[b]{.45\textwidth}
         \centering
         \includegraphics[width=\textwidth]{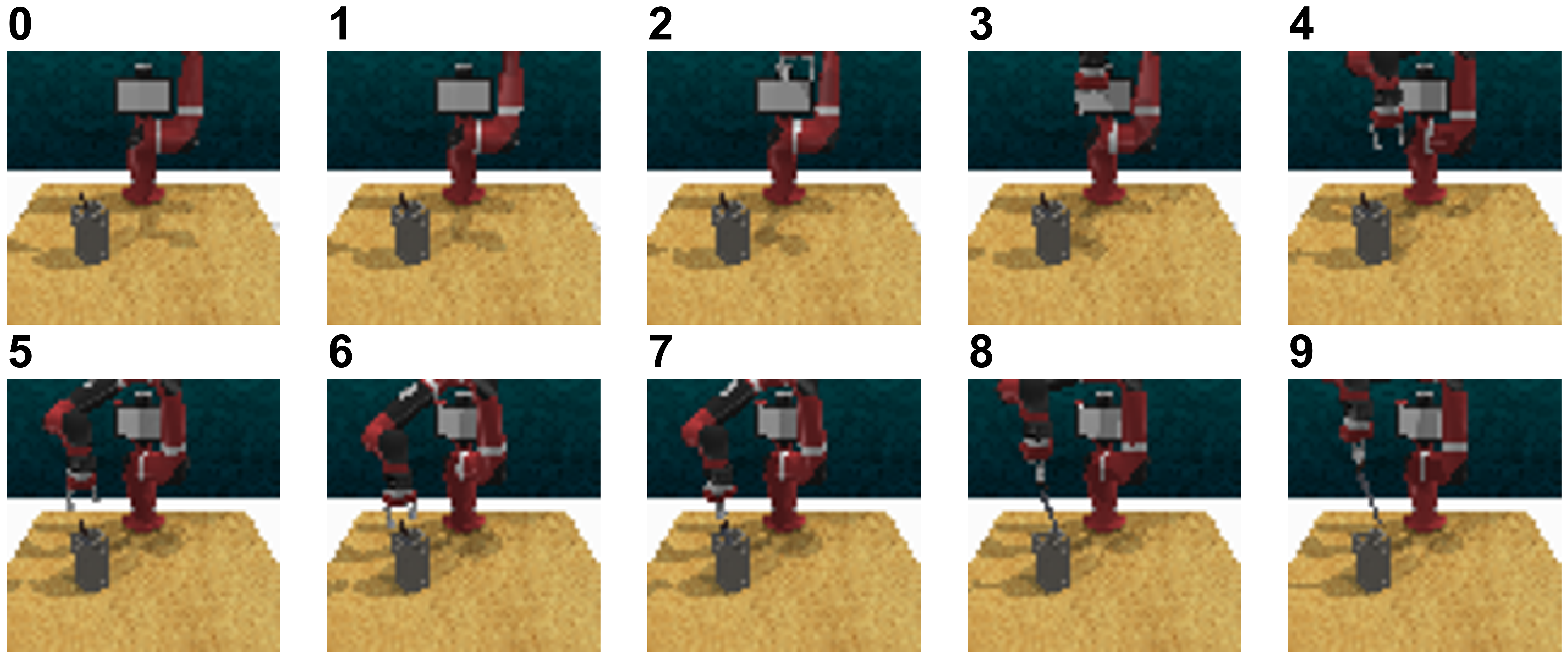}
         \includegraphics[width=\textwidth]{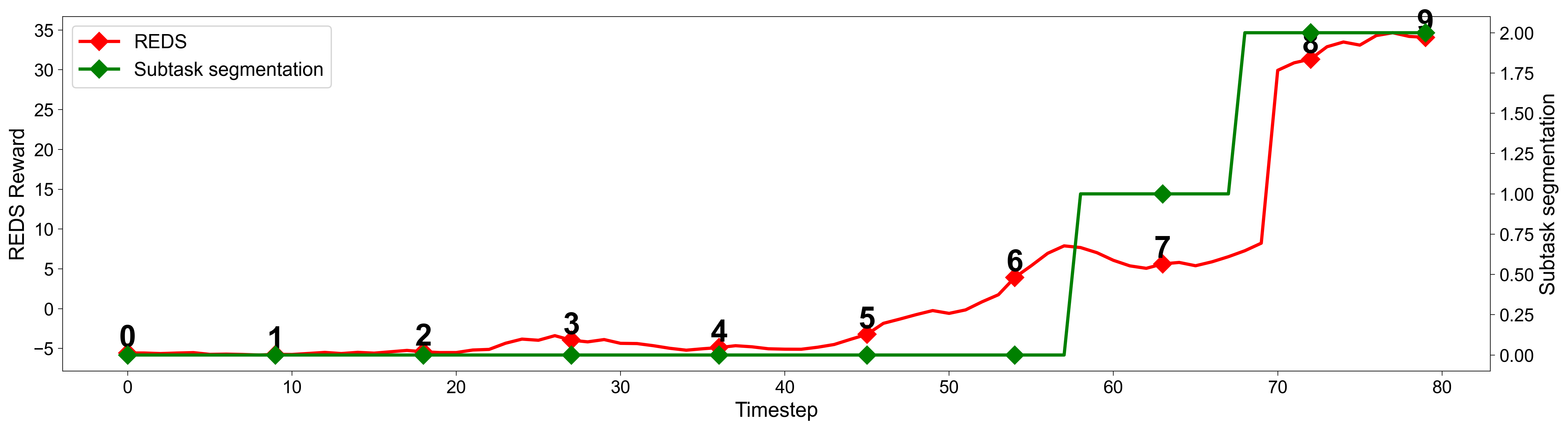}
         \label{fig:reds_rlbench_sawyer}
         \vspace{-0.15in}
         \caption{(Unseen) Sawyer}
    \end{subfigure}
    \\
    \caption{
    Qualitative results of REDS with different robot embodiments. REDS was trained using demonstrations from the Panda Arm and evaluated on an unseen demonstration from the Sawyer Arm in Take Umbrella Out of Stand from RLBench~\citep{james2020rlbench}. We visualize several frames above the graph and mark them with a diamond symbol. 
    }
    \label{fig:qual_embodiment}
    \vspace{-0.05in}
\end{figure}

        \paragraph{Transfer to unseen embodiments} Since our framework leverages only action-free video data, we hypothesize that transferring to other robot embodiments with similar DoFs is feasible. To support this claim, we train \metabbr with demonstrations of the Franka Panda Arm and then compute the reward of an unseen demonstration of the Sawyer Arm in Take Umbrella Out of Stand from RLBench~\citep{james2020rlbench}. Figure~\ref{fig:qual_embodiment} shows that \metabbr generates informative reward signals even with the unseen embodiment. For instance, \metabbr can capture the behavior of taking the umbrella out of the stand, as indicated by the increased reward signals between 6 and 7. Additionally, Figure~\ref{fig:embodiment_graph} shows that REDS trained only with the Panda Arm can be used to train downstream RL agents in the environment with the Sawyer Arm. 
        \vspace{-0.06in}
 
        \begin{figure}[t]
    \centering
    \begin{subfigure}[b]{0.24\textwidth}
        \centering
        \includegraphics[width=\textwidth]{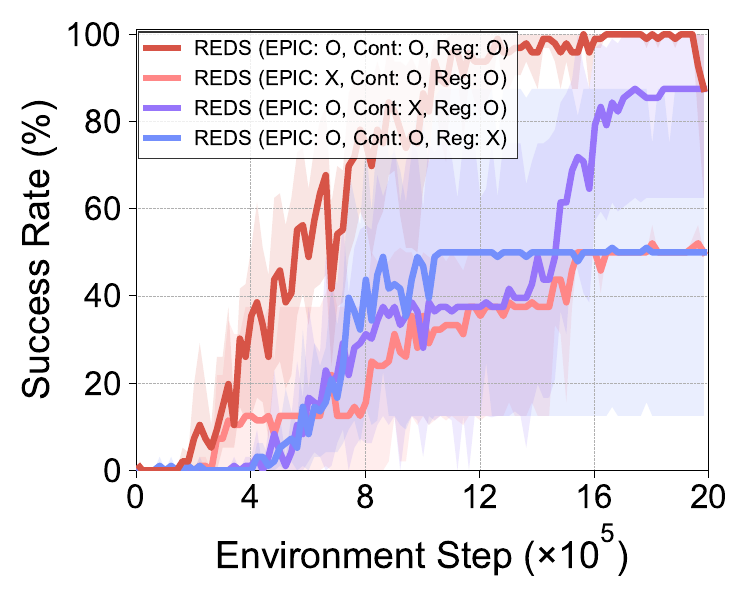}
        \vspace{-0.22in}
        \caption{Training objectives}
        \label{fig:mw_abl_training_obj}
    \end{subfigure}
    \begin{subfigure}[b]{0.24\textwidth}  
        \centering 
        \includegraphics[width=\textwidth]{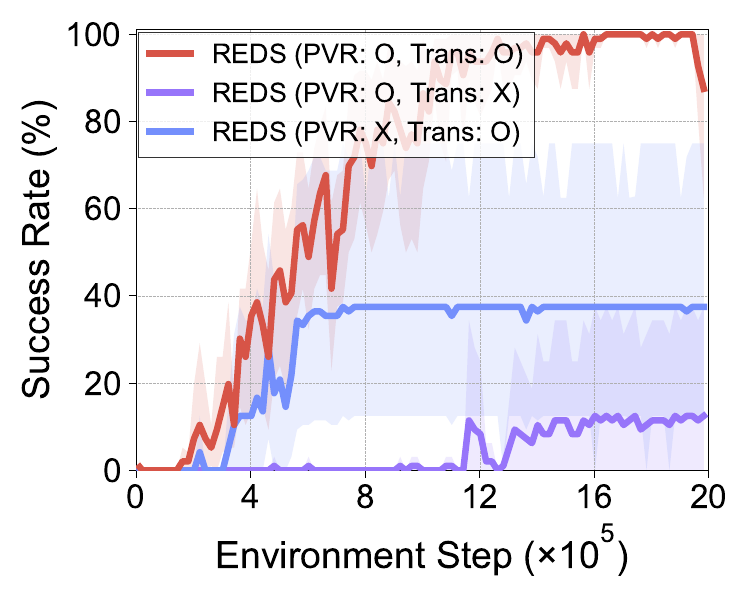}
        \vspace{-0.22in}
        \caption{Architecture}
        \label{fig:mw_abl_arch}
    \end{subfigure}
    \begin{subfigure}[b]{0.24\textwidth}   
        \centering 
        \includegraphics[width=\textwidth]{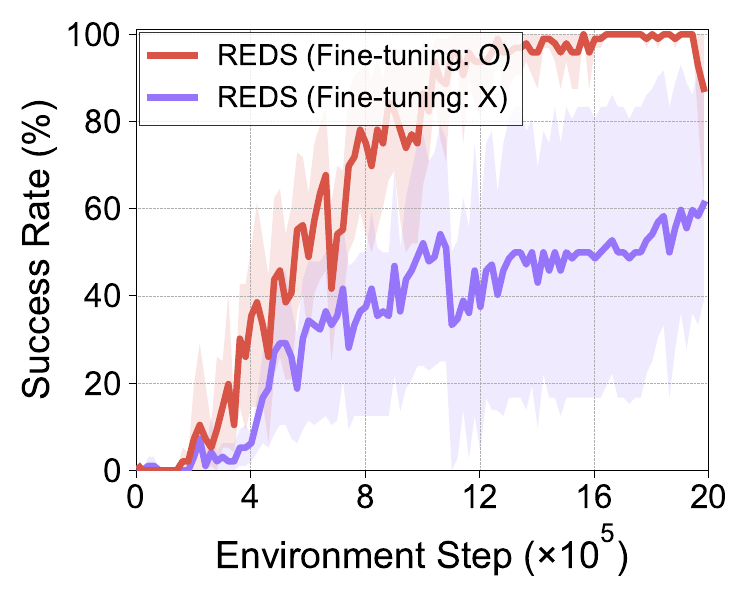}
        \vspace{-0.22in}
        \caption{Fine-tuning}
        \label{fig:mw_abl_refinement}
    \end{subfigure}
    \begin{subfigure}[b]{0.24\textwidth}   
        \centering 
        \includegraphics[width=\textwidth]{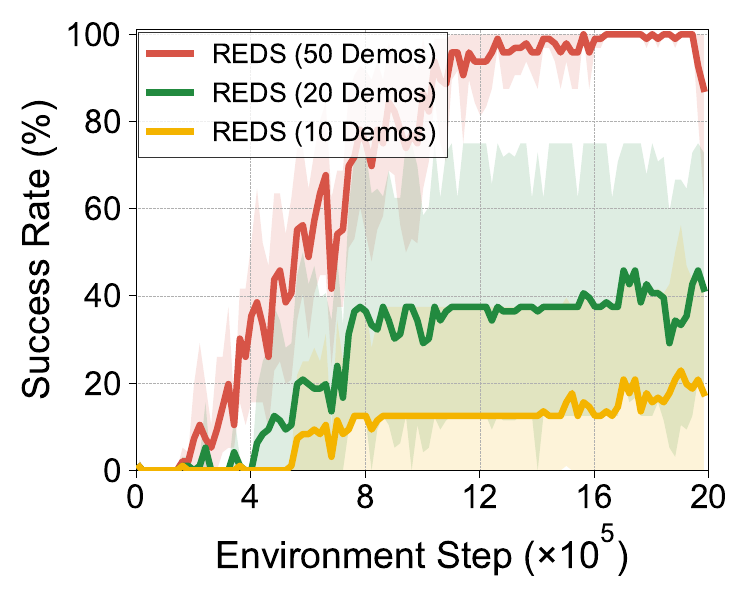}
        \vspace{-0.22in}
        \caption{Expert demonstrations}
        \label{fig:mw_abl_demo}
    \end{subfigure}

    \caption{
    Learning curves for two Meta-world~\citep{yu2020meta} robotic manipulation tasks, measured by success rate (\%), to examine the effects of (a) training objectives, (b) architecture, (c) fine-tuning, and (d) the number of expert demonstrations. The solid line and shaded regions show the mean and stratified bootstrap interval across 8 runs.
    }
    \vspace{-0.2in}
    \label{fig:ablation}
\end{figure}
    \subsection{Ablation Studies}
    \label{exp:abl}
        \paragraph{Effect of training objectives} We investigate the effect of each training objective in Figure~\ref{fig:mw_abl_training_obj}. Specifically, we compare \metabbr with 1) a baseline trained with regression to subtask segmentation instead of EPIC loss $\mathcal{L}_{\text{EPIC}}$, 2) a baseline that utilizes only video representations without subtask embeddings, and 3) a baseline trained without the regularization loss $\mathcal{L}_{\text{reg}}$. We observe that RL performance significantly degrades without each component, implying that our losses synergistically improve reward quality. 
        
        \vspace{-0.1in}
        \paragraph{Effect of architecture} To verify the design choice proposed in Section~\ref{sec:arch}, we compare \metabbr with 1) a baseline using a CNN for encoding images instead of pre-trained visual representations (PVR) and 2) a baseline simply concatenating pre-trained visual representations without a causal transformer. Figure~\ref{fig:mw_abl_arch} shows that both baselines show worse performance compared to ours. Notably, detaching a causal transformer significantly degrades RL performance, implying that temporal information is essential for providing suitable reward signals in robotic manipulation.

        \vspace{-0.1in}
        \paragraph{Effect of fine-tuning} In Figure~\ref{fig:mw_abl_refinement}, we compare \metabbr trained only with the expert demonstrations in the initial phase to \metabbr fine-tuned with additional suboptimal demonstrations as described in Section~\ref{sec:train_infer}. \metabbr shows improved RL performance when trained with additional suboptimal demonstrations, indicating that the coverage of state distribution impacts the reward quality. Further investigation on how to efficiently collect suboptimal demonstrations to enhance the performance of learned reward function is a promising future direction.
        
        \vspace{-0.1in}
        \paragraph{Effect of the number of expert demonstrations} We investigate the effect of the number of expert demonstrations by measuring the RL performance of DreamerV3 agents with \metabbr trained with different numbers of expert demonstrations in 2 tasks (Door Open, Drawer Open) from Meta-world. Figure~\ref{fig:mw_abl_demo} shows that the agents' RL performance positively correlates with the number of expert demonstrations trained for reward learning.
        \vspace{-0.1in}

\section{Conclusion}
\label{sec:conclusion}
    \vspace{-0.07in}
    We proposed \metabbr, a visual reward learning framework considering subtasks by utilizing subtask segmentation. Our main contribution is based on proposing a new reward model leveraging minimal domain knowledge as a ground-truth reward function. Our approach is generally applicable and does not require any additional instrumentations in online interactions. We believe \metabbr will significantly alleviate the burden of reward engineering and facilitate the application of RL to a broader range of real-world robotic tasks.

\section*{Acknowledgements}
    This research is supported by Institute of Information \& Communications Technology Planning \& Evaluation (IITP) grant funded by the Korea government (MSIT) (RS-2022-II220953, Self-directed AI Agents with Problem-solving Capability; RS-2019-II190075 Artificial Intelligence Graduate School Program (KAIST); No. RS-2024-00509279, Global AI Frontier Lab).

\section*{Limitation and Future Directions}
    One limitation of our work is that we assume the knowledge of the object-centric subtasks in a task. For automating subtask definition and segmentation in new tasks and new domains other than robotic manipulations, investigating the planning and reasoning capabilities of pre-trained Multimodal Large Language Model (MLLM)~\citep{park2023generative, honerkamp2024language, zawalskirobotic, shah2024bumble, liu2024moka} would be an intriguing research direction.
    
    Additionally, the performance of REDS relies on pre-trained representations trained with natural image/text data for encoding videos and subtask instructions. Although \metabbr proves its effectiveness in various robotic manipulation tasks, we observe that \metabbr struggles to distinguish subtle changes (e.g., screwing the leg in One Leg) even with pre-trained representations trained on ego-centric motion videos~\citep{ma2023liv}. We believe that the quality of rewards can be further improved by utilizing 1) pre-trained representations with large-scale data with diverse robotic tasks~\citep{padalkar2023open, khazatsky2024droid} and 2) representations trained with objectives considering affordances~\citep{bahl2023affordances} or object-centric methods~\citep{devin2018deep}.
    
    Furthermore, there is room for improvement to enhance generalization and robustness. Although our experiments are designed to evaluate generalization in unseen environments, they may face challenges in out-of-distribution environments, such as significant changes in the background or camera angles. Future work could address these challenges through data augmentation or domain adaptation techniques. while our contrastive learning objective currently focuses on minimizing distances between relevant video and text embeddings, the reward model may generate inappropriate reward signals for semantically different subtasks that share similar video content and text instructions. Incorporating a loss term to maximize distances between irrelevant embeddings could further improve robustness in tasks with similar subtasks, and we will explore this enhancement in future work.

    Finally, the number of expert demonstrations and the number of iterations for fine-tuning \metabbr are determined by empirical trials. Investigating how to collect failure demonstrations to mitigate reward misspecification efficiently is an interesting future direction.
    
\section*{Ethic Statement}
    Video demonstrations and subtask segmentations used in the experiments were sourced from publicly available benchmarks (Meta-world, RLBench, FurnitureBench), ensuring no personal or sensitive information is involved. Potential risks could arise when training and deploying RL agents directly in real-world scenarios, particularly in human-robot interactions. Ensuring the safety and reliability of these agents before deployment is essential to prevent harm.

\section*{Reproducibility Statement}
    For the reproducibility of \metabbr, we have provided a detailed explanation of implementation details and experimental setups in Section~\ref{sec:train_infer}, Section~\ref{sec:exp}, and Appendix~\ref{appendix:impl}. In addition, to further facilitate the reproduction, we release the open-sourced implementation through the \href{https://csmile-1006.github.io/reds/}{project website}.


\bibliography{iclr2025_conference}  
\bibliographystyle{iclr2025_conference}

\newpage
\appendix

\section{Experiment Details}
\label{appendix:impl}
    \paragraph{Training and inference details} To ensure robustness against visual changes, we apply data augmentations, including random shifting~\citep{yarats2021image, yarats2022mastering} and color jittering. For optimization, we train \metabbr with AdamW~\citep{loshchilov2018decoupled} optimizer with a learning rate of $1 \times 10^{-4}$, weight decay of $2 \times 10^{-2}$, and a cosine decay schedule for adjusting the training learning rate. We apply a warm-up scheduling for the initial 500 gradient steps starting from a learning rate of 0. Note that the parameters for CLIP visual/text encoders have not been updated. For training downstream RL agents, we normalize the reward by dividing it by the maximum value observed in the expert demonstrations. We report the hyperparameters used in our experiments in Table~\ref{tbl:hp}. For both coverage distribution $\mathcal{D}_{\mathcal{C}}$ and potential shaping distribution $\mathcal{D}_\mathcal{S}$, we use the same dataset with subtask segmentations $\mathcal{D}^{i}$, unlike prior work dealing with arbitrarily random distributions because of the absence of subtask segmentations. To canonicalize our reward functions, we estimate the expectation over state distributions using a sample-based average over 8 additional samples from $\mathcal{D}$ per sample. To prevent false positive cases in predicting subtasks in online interactions, we add margins to similarity scores inversely proportional to the subtasks in online interactions. Specifically, we infer the subtask $\hat{i}$ as follows:
    \begin{equation}
        \hat{i} = {\mathop{\mathrm{argmax}}}_{i \in \{1, ..., k\}}(\mathrm{sim}(\bm{v}_t, \bm{e}_{i}) + \eta * (k - i)),
    \end{equation}
    where $\eta$ is a hyperparameter for the margin between subtasks. For each subtask $U_i$, we compute similarity scores between the visual observations within the subtask from expert demonstrations and their corresponding instructions. The threshold ${T_{U_i}}$ is set to the 75th percentile of these scores to account for demonstration variability while capturing the most relevant matches. Please refer to Figure~\ref{fig:mw_abl_fail_threshold} for supporting experiments.
    \begin{table}[ht]
\caption{Hyperparameters of \metabbr used in our experiments.}
\label{tbl:hp}
\begin{center}
\small\resizebox{.8\textwidth}{!}{
\begin{tabular}{l l}
\toprule
\textbf{Hyperparameter} & \textbf{Value} \\
\midrule
Batch size & 32 (Meta-world, RLBench), 8 (FurnitureBench) \\
Training steps & 5000 \\
Learning rate & 0.0001  \\
Optimizer & AdamW \citep{loshchilov2018decoupled}\\
Optimizer momentum & $\beta_1 = 0.9$, $\beta_2 = 0.999$ \\
Weight decay & 0.02 \\
Learning rate decay & Linear warmup and cosine decay \\
Warmup steps & 500 \\
Context length & 4 \\
Causal transformer size & 3 layers, 8 heads, 512 units\\
\midrule
EPIC canonical samples & 8 \\
$\epsilon$ for progressive reward signal & 0.05 \\
$\eta$ for inferring subtasks & 0.01 (Meta-world, RLBench), 0.05 (FurnitureBench) \\
number of training iterations $n$ & 2 (Meta-world, RLBench), 1 (FurnitureBench) \\
\bottomrule
\end{tabular}
}
\end{center}
\end{table} 
    
    \paragraph{Meta-world experiments} We use visual observations of $64 \times 64 \times 3$. To consistently use a single camera viewpoint over all tasks, we use the modified version of the $\mathtt{corner2}$ viewpoint as suggested by \citet{seo2023masked}. Expert demonstrations for each task are collected using scripted policies publicly released in the benchmark. We use an action repeat of 2 to accelerate training and set the maximal episode length as 250 for all Meta-world tasks. For downstream RL, we use the implementation of DreamerV3 from VIPER\footnote{\url{https://github.com/Alescontrela/viper_rl}}. We report the hyperparameters of DreamerV3 agents used in our experiments in Table~\ref{tbl:dreamer_hp}. Unless otherwise specified, we use the same set of hyperparameters as VIPER. For all ablation experiments (Figure~\ref{fig:visual_distraction},~\ref{fig:ablation},~\ref{fig:additional_ablation}), we report results in Door Open and Drawer Open.
    \paragraph{RLBench experiments} For training both reward models and downstream RL agents, we utilize $64 \times 64 \times 3$ RGB observations from the front camera and wrist camera. For downstream RL, we don't use any expert demonstrations, and we use the same set of hyperparameters as VIPER.
    
    \paragraph{FurnitureBench experiments} We use the implementation of IQL from FurnitureBench \footnote{\url{https://github.com/clvrai/furniture-bench/tree/main/implicit_q_learning}} for our experiments. We utilize $224 \times 224 \times 3$ RGB observations from the front camera and wrist cameras, along with proprioceptive states, to represent the current state. We encode each image with pre-trained R3M~\citep{nair2022rm} for visual observations. Following~\cite {kostrikov2022offline}, we first run offline RL for 1M gradient steps, then continue training while collecting environment interaction data, adding it to the replay buffer, and repeating this process for 150 episodes. Before online fine-tuning, we pre-fill the replay buffer with 10 rollouts from the pre-trained IQL policy. We adopt techniques from RLPD~\citep{ball2023efficient} for efficient offline-to-online RL training. Specifically, we sample 50\% of the data from the replay buffer and the remaining 50\% from the offline data buffer containing 300 expert demonstrations. We also apply LayerNorm~\citep{ba2016layer} in the critic/value network of the IQL agent to prevent catastrophic overestimation. We list the hyperparameters used in our experiments in Table~\ref{tbl:furniture_hp}. For training REDS, we collect subtask segmentations for suboptimal demonstrations using the automatic subtask inference procedure described in Section~\ref{sec:train_infer}, and we manually modified some subtask segmentations with false negatives to guarantee stable performance.
    \begin{figure*}[t]
\begin{minipage}[h!]{.48\textwidth}
\captionof{table}{
    Hyperparameters of DreamerV3~\citep{VIPER} used in Meta-world experiments.
}
\label{tbl:dreamer_hp}
\centering \small
\resizebox{.85\textwidth}{!}{
\begin{tabular}{l l}
\toprule
\textbf{Hyperparameter} & \textbf{Value} \\

\midrule
General & \\
\midrule
Replay Capacity (FIFO) & $5 \times 10^{5}$ \\
Start learning (prefill) & 5000 \\
MLP size & $2 \times 512$ \\
\midrule
World Model & \\
\midrule
RSSM size & 512 \\
Base CNN channels & 32 \\
Codes per latent & 32 \\

\bottomrule
\end{tabular}
}

\end{minipage}
~~~
\begin{minipage}[h!]{.48\textwidth}
    \captionof{table}{Hyperparameters of IQL~\citep{kostrikov2022offline} used in FurnitureBench experiments.}
    \label{tbl:furniture_hp}
    \centering\small\resizebox{\textwidth}{!}{
        \begin{tabular}{l l}
        \toprule
        \textbf{Hyperparameter} & \textbf{Value} \\
        \midrule
        Learning rate & $3 \times 10^{-4}$ \\
        Batch size & 256 \\
        Policy \# hidden units & (512, 256, 256) \\
        Critic/value \# hidden units & (512, 256, 256) \\
        Image encoder & R3M~\citep{nair2022rm} \\
        Discount factor ($\gamma$) & 0.996 \\
        Expectile ($\tau$) & 0.8 \\
        Inverse Temperature ($\beta$) & 10.0 \\
        \bottomrule
        \end{tabular}
    }
\end{minipage}
\end{figure*}
    \paragraph{Computation} We use 24 Intel Xeon CPU @ 2.2GHz CPU cores and 4 NVIDIA RTX 3090 GPUs for training our reward model, which takes about 1.5 hours in Meta-world and 3 hours in FurnitureBench due to high-resolution visual observations from multiple views. For training DreamerV3 agents in Meta-world, we use 24 Intel Xeon CPU @ 2.2GHz CPU cores and a single NVIDIA RTX 3090 GPU, which takes approximately 4 hours over 500K environment steps. For training IQL agents in FurnitureBench, we use 24 Intel Xeon CPU @ 2.2GHz CPU cores and a single NVIDIA RTX 3090 GPU, taking approximately 2 hours for 1M gradient steps in offline RL and 4.5 hours over 150 episodes of environment interactions in online RL.

\section{\metabbr Architecture Details}
    We encode visual observations with a pre-trained CLIP~\citep{radford2021learning} ViT-B/16 visual encoder, utilizing all representations from the sequence of patches. We adopt 1D learnable parameters with the same size for positional embedding, and we add these parameters to 2D fixed sin-cos embeddings and add them to features. To encode temporal dependencies in visual observations, we use a GPT~\citep{radford2018improving} architecture with 3 layers and 8 heads. In FurnitureBench, we use a sequence of images from both the front camera and wrist camera as input. Given $s^{\mathrm{front}}_{t}/s^{\mathrm{wrist}}_{t}$ from the front/wrist camera, we concatenate visual observations to $[o^{\mathrm{front}}_{t-K-1}, o^{\mathrm{wrist}}_{t-K-1}, ..., o^{\mathrm{front}}_{t}, o^{\mathrm{wrist}}_t]$, add positional embeddings, 2D fixed sin-cos embeddings, and additional 1D learnable parameters for each viewpoint for effectively utilizing images from multiple cameras. We then pass the features to the transformer layer, the same as the model with a single image. The subtask embedder and final reward predictor are implemented as 2-layer MLPs. 
    
\section{Baseline Details}
\label{appendix:baseline}
    \paragraph{ORIL~\citep{oril}} For implementing ORIL with visual observations, we use the CNN architecture from \citet{yarats2021image} to encode image observations. For training data, we use the same set of demonstrations as for training \metabbr. Since our training data are divided into success and failure demonstrations, we do not use positive-unlabeled learning~\citep{xu2021positive} in our experiments. For robustness against visual changes, we apply the same augmentation techniques used for training \metabbr. 
    
    \paragraph{Rank2Reward (R2R)~\citep{r2r}} To ensure compatibility with backbone RL algorithms~\citep{hafner2023mastering, kostrikov2022offline} implemented in JAX, we reimplement the reward model with JAX following the official implementation of Rank2Reward~\footnote{\url{https://github.com/dxyang/rank2reward}} and use the same hyperparameters. We first pre-train the ranking network using the same expert demonstrations as \metabbr, and we then train a discriminator for the expert demonstration and policy rollouts, weighted by the output from the pre-trained ranking network. For training efficiency, we use the CNN architecture from \citet{yarats2021image} for encoding visual observations instead of R3M~\citep{nair2022rm}, finding no significant difference when we use the pre-trained visual representations like R3M, but with much slower training in online RL. We observe that our R2R implementation with DreamerV3 in JAX outperforms the original version implemented with DrQ-V2~\citep{yarats2022mastering} agents.
    
    \paragraph{DrS~\citep{drs}} Similar to R2R, we reimplement DrS with JAX following the official implementation of DrS~\footnote{\url{https://github.com/tongzhoumu/DrS}}, and use the same set of hyperparameters for reward learning. As the original DrS implementation is based on a state-based environment, we switch the backbone RL algorithm from SAC to DrQ-V2~\citep{yarats2022mastering} and apply the augmentation technique in the reward learning phase for processing visual observations efficiently. To report the RL performance, we use the learned dense reward model to train new RL agents. In FurnitureBench experiment, we train the reward model with the same expert/failure demonstrations as in Section~\ref{exp:fb}, without online interaction, to avoid unsafe behaviors and a significant increase in training time from online interactions.
    
    \paragraph{VIPER~\citep{VIPER}} We use the official implementation of VIPER~\footnote{\url{https://github.com/Alescontrela/viper_rl}} for our experiments. Given the similarities among robotic manipulation tasks, we use the same set of hyperparameters as in RLBench~\citep{james2020rlbench} experiments to train VQ-GAN and VideoGPT. We train 100K steps, choosing the checkpoint with the minimum validation loss. In FurnitureBench experiment, we use images from the front camera, resized to $64 \times 64 \times 3$, and set the exploration objective $\beta$ as 0.

\section{Task Descriptions}
\label{appendix:task}
In this section, we list the subtasks and corresponding text instructions for each task in Table~\ref{tbl:task}. For Meta-world tasks, we provide the code snippet used to determine the success of each subtask (Please refer to the Meta-world~\citep{yu2020meta} for more details). For the FurnitureBench One Leg task, we outline the criteria used by human experts to assess the success of each subtask based on the metric defined in FurnitureBench~\citep{furniturebench}.
\begin{table}[ht]
\caption{A list of subtasks and language description for each subtask used for \metabbr in our experiments.}
\label{tbl:task}
\begin{center}
\resizebox{\textwidth}{!}{
    \begin{tabular}{@{}llll@{}}
        \toprule
        Task                       & Subtask & Success condition                                             & Language description                                                                      \\ \midrule
        Meta-world Faucet Close    & 1       & $\mathrm{object\_grasped} \leq 0.9$                                  & a robot arm reaching the faucet handle.                                          \\
                                   & 2       & $\mathrm{target\_to\_obj} \leq 0.07$                             & a robot arm rotating the faucet handle to the right.                             \\ \midrule
        Meta-world Drawer Open     & 1       & $\mathrm{gripper\_error} \leq 0.03$                              & a robot arm grabbing the drawer handle.                                          \\
                                   & 2       & $\mathrm{handle\_error} \leq 0.03$                                     & a robot arm opening a drawer to the green target point.                          \\
                                   & 3       & $\mathrm{handle\_error} \leq 0.03$                               & a robot arm holding the drawer handle near the green target point after opening. \\ \midrule
        Meta-world Lever Pull      & 1       & $\mathrm{ready\_to\_lift} > 0.9$                             & a robot arm touching the lever.                                                  \\
                                   & 2       & $\mathrm{lever\_error} \leq \mathrm{np.pi / 24}$                          & a robot arm pulling up the lever to the red target point.                        \\ \midrule
        Meta-world Door Open       & 1       & $\mathrm{reward\_ready} \geq 1.0$                             & a robot arm grabbing the door handle.                                            \\
                                   & 2       & $\mathrm{abs(obs{[}4{]}} - \mathrm{self.\_target\_pos{[}0{]}}) \leq 0.08$ & a robot arm opening a door to the green target point.                            \\
                                   & 3       & $\mathrm{abs(obs{[}4{]}} - \mathrm{self.\_target\_pos{[}0{]}}) \leq 0.08$       & a robot arm holding the door handle near the green target point after opening.   \\ \midrule
        Meta-world Coffee Pull     & 1       & $\mathrm{tcp\_to\_obj} < 0.04 \wedge \mathrm{tcp\_open} > 0$      & a robot arm grabbing the coffee cup.                                             \\
                                   & 2       & $\mathrm{obj\_to\_target} \leq 0.07$                             & a robot arm moving the coffee cup to the green target point.                     \\
                                   & 3       & $\mathrm{obj\_to\_target} \leq 0.07$                                   & a robot arm holding the cup near the green target point.                         \\ \midrule
        Meta-world Peg Insert Side & 1       & $\mathrm{tcp\_to\_obj} < 0.03 \wedge \mathrm{tcp\_open} > 0$      & a robot arm grabbing the green peg.                                              \\
                                   & 2       & $\mathrm{obj{[}2{]}} - 0.1 > \mathrm{self.obj\_init\_pos{[}2{]}}$      & a robot arm lifting the green peg from the floor.                                \\
                                   & 3       & $\mathrm{obj\_to\_target} \leq 0.07$                             & a robot arm inserting the green peg to the hole of the red box.                  \\
                                   & 4       & $\mathrm{obj\_to\_target} \leq 0.07$                                   & a robot arm holding the green peg after inserting.                               \\ \midrule
        Meta-world Push            & 1       & $\mathrm{tcp\_to\_obj} \leq 0.03$                                & a robot arm grabbing the red cube.                                               \\
                                   & 2       & $\mathrm{target\_to\_obj} \leq 0.05$                             & a robot arm pushing the grabbed red cube to the green target point.              \\
                                   & 3       & $\mathrm{target\_to\_obj} \leq 0.05$                                   & a robot arm holding the grabbed red cube near the green target point.            \\ \midrule
        Meta-world Sweep Into      & 1       & $\mathrm{self.touching\_main\_object > 0} \wedge \mathrm{tcp\_opened} > 0$    & a robot arm grabbing the red cube.                                               \\
                                   & 2       & $\mathrm{target\_to\_obj} \leq 0.05$                             & a robot arm sweeping the grabbed red cube to the blue target point.              \\
                                   & 3       & $\mathrm{target\_to\_obj} \leq 0.05$                                   & a robot arm holding the grabbed red cube near the blue target point.             \\ \midrule
        Meta-world Door Close      & 1       & $\mathrm{in\_place} == 1.0$                                              & a robot arm grabbing the door handle.                                            \\
                                   & 2       & $\mathrm{obj\_to\_target} \leq 0.08$                             & a robot arm closing a door to the green target point.                            \\
                                   & 3       & $\mathrm{obj\_to\_target} \leq 0.08$                                   & a robot arm holding the door handle near the green target point after closing.   \\ \midrule
        Meta-world Window Close    & 1       & $\mathrm{tcp\_to\_obj} \leq 0.05$                                & a robot arm grabbing the window handle.                                          \\
                                   & 2       & $\mathrm{target\_to\_obj} \leq 0.05$                             & a robot arm closing a window from left to right.                                 \\
                                   & 3       & $\mathrm{target\_to\_obj} \leq 0.05$                                   & a robot arm holding the window handle after closing.                             \\ \midrule
        FurnitureBench One Leg     & 1       & robot gripper tips make contact with one surface of the tabletop. & a robot arm picking up the white tabletop.                                       \\
                                   & 2       & nearest corner of the tabletop is placed close to the right edge of the obstacle. & a robot arm pushing the white tabletop to the front right corner.                \\
                                   & 3       & robot gripper securely grasps a leg of the table and lifts it. & a robot arm picking up the white leg.                                            \\
                                   & 4       & leg is inserted into one of the screw holes of the tabletop, and the robot releases the gripper. & a robot arm inserting the white leg into screw hole.                             \\
                                   & 5 & leg is fully assembled to the tabletop.  & a robot arm screwing the white leg until tightly lifted.                         \\
                                   & 6       & leg is fully assembled to the tabletop.  & a robot arm holding the white leg in place.                                      \\ \midrule
        RLBench Take Umbrella Out of Umbrella Stand & 1 & $\mathrm{GraspedCondition(self.robot.gripper, self.umbrella).condition_met()[0]}$  & a robot arm grasping the umbrella.    \\
                                   & 2       & $\mathrm{DetectedCondition(self.umbrella, self.success_sensor, negated=True).condition_met()[0]}$ & a robot arm taking the grasped umbrella ouf of the umbrella stand.                \\
                                   & 3       & $\mathrm{DetectedCondition(self.umbrella, self.success_sensor, negated=True).condition_met()[0]}$ & a robot arm holding the umbrella on the umbrella stand.                                            \\ \bottomrule
    \end{tabular}
}
\end{center}
\end{table} 
\begin{table}[ht]
\caption{A list of language description used for CLIP and LIV.}
\label{tbl:task_description_clip}
\begin{center}
\resizebox{\textwidth}{!}{
    \begin{tabular}{@{}ll@{}}
        \toprule
        Task                   & Language description                                                                 \\ \midrule
        Meta-world Door Open   & a robot arm grabbing the drawer handle and opening the drawer.                       \\
        Meta-world Drawer Open & a robot arm grabbing the door handle and opening the door to the green target point. \\ \bottomrule
    \end{tabular}
}
\end{center}
\end{table} 
\section{Extended Related Work}
\label{appendix:related_work}
    \paragraph{Quantifying differences between reward functions} Previous work has explored methods for measuring the difference between reward functions without relying on policy optimization procedures~\citep{EPIC, wulfe2022dynamicsaware, skalse2024starc}. In particular, \citet{EPIC} introduced the EPIC distance, a pseudometric invariant to equivalent classes of reward functions. Subsequent work~\citep{rocamonde2024visionlanguage, adeniji2022skill, liang2022reward} has employed EPIC to assess the quality of reward functions. In this paper, we take a different approach by using EPIC distance as an optimization objective. While \citet{adeniji2022skill} also utilizes EPIC distance for optimizing intrinsic reward functions in skill discovery, our method applies EPIC distance to train dense reward functions for long-horizon tasks, serving as a direct reward signal for RL training.

    \paragraph{Segmenting demonstrations for long-horizon manipulation tasks} Several approaches have been proposed to decompose long-horizon demonstrations into multiple subgoals to prevent error accumulation and provide intermediate signals for agent training. These include extracting key points from proprioceptive states~\citep{james2022q, james2022coarse, shridhar2023perceiver, shi2023shiwaypoint}, employing greedy heuristics on off-the-shelf visual representations pre-trained with robotic data~\citep{zhang2024universal}, and learning additional modules on top of pre-trained visual-language models to align with keyframes~\citep{koukisa}. Our work builds on these efforts by leveraging subtask segmentations but focuses on developing a reward learning framework that explicitly incorporates subtask decomposition to generate suitable reward signals for intermediate tasks. Additionally, we further demonstrate that our model generalizes effectively to unseen tasks and robot embodiments.

\clearpage
\section{Extended Qualitative Analysis}
\label{appendix:qualitative}
     \begin{figure} [htbp] \centering
    \begin{subfigure}[b]{.48\textwidth}
         \centering
         \includegraphics[width=\textwidth]{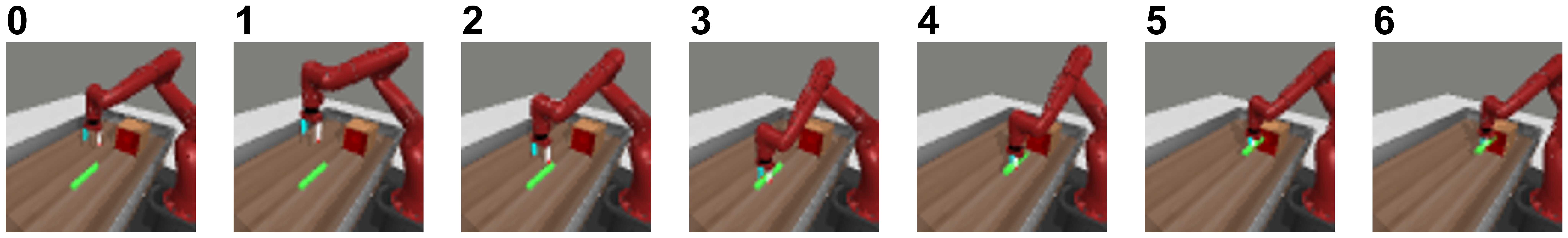}
         \includegraphics[width=\textwidth]{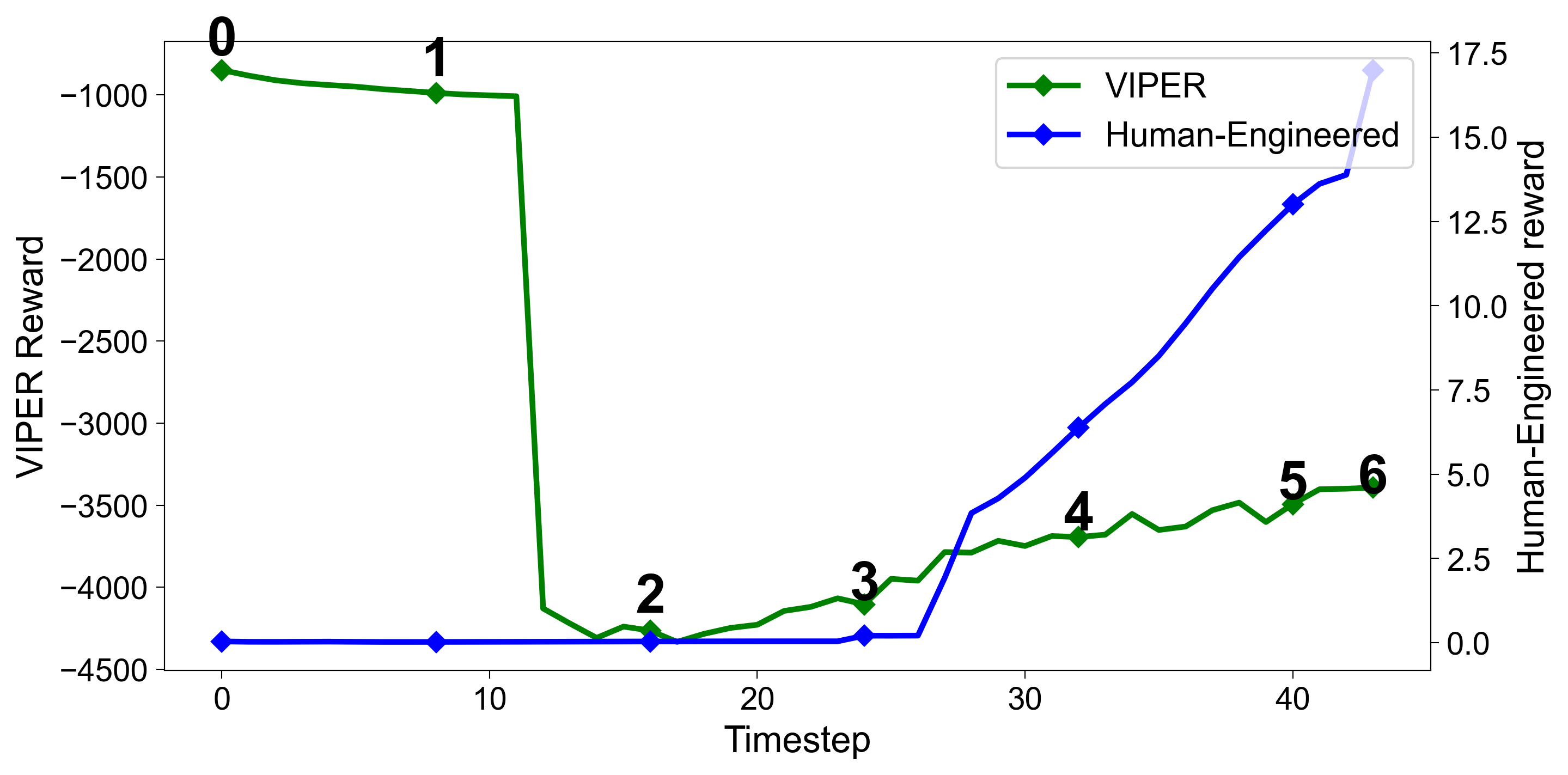}
         \includegraphics[width=\textwidth]{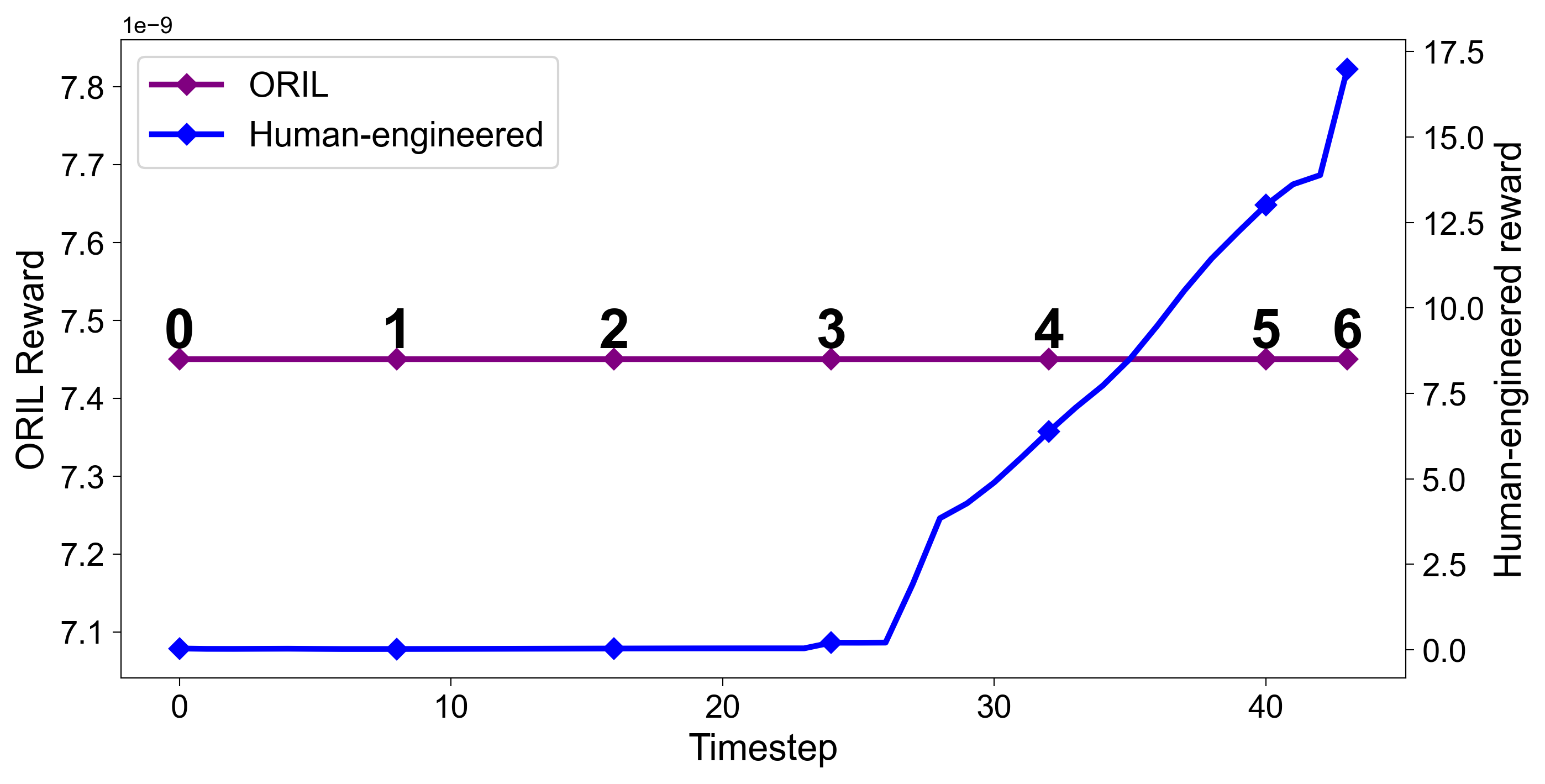}
         \includegraphics[width=\textwidth]{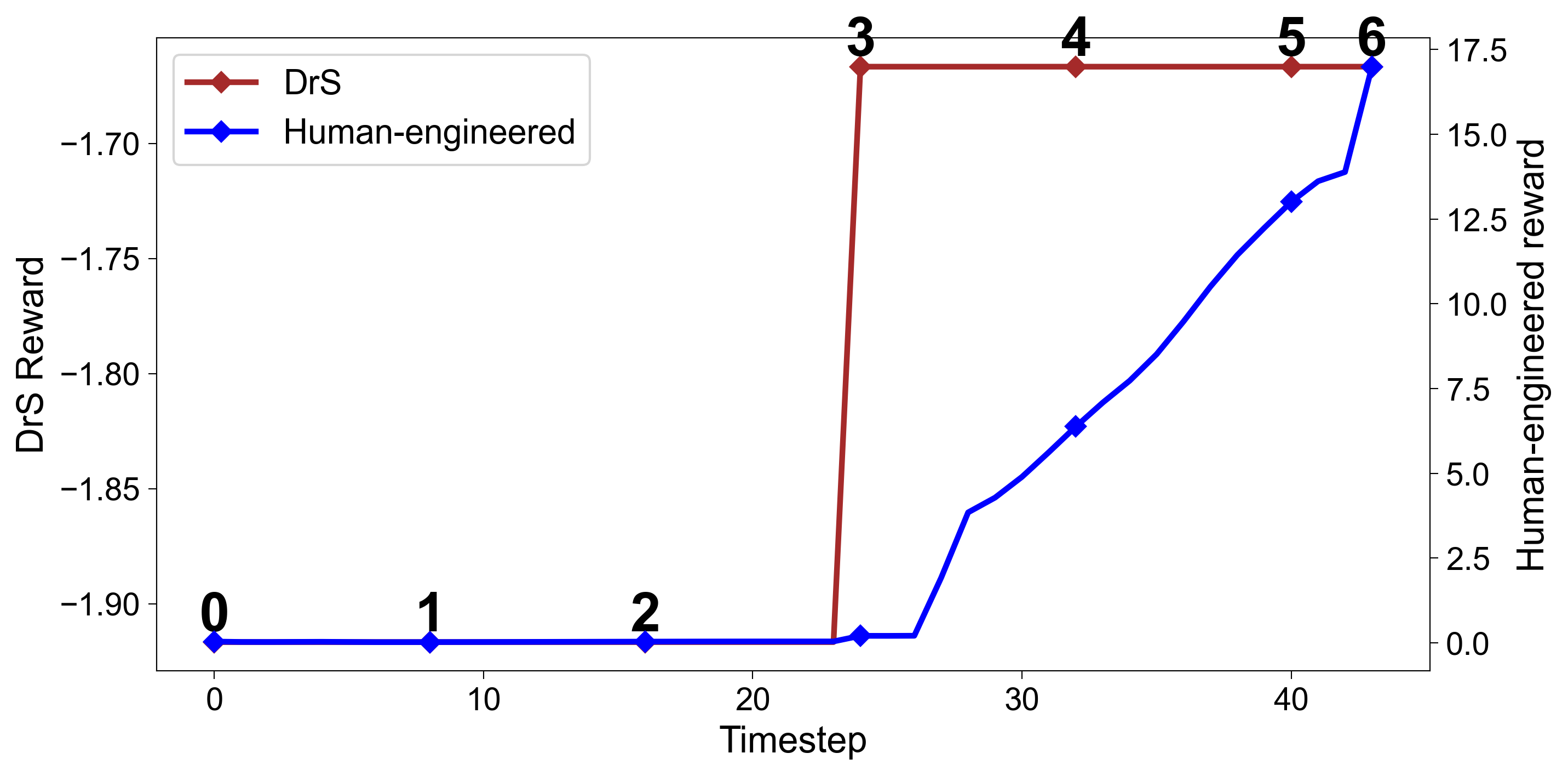}
         \includegraphics[width=\textwidth]{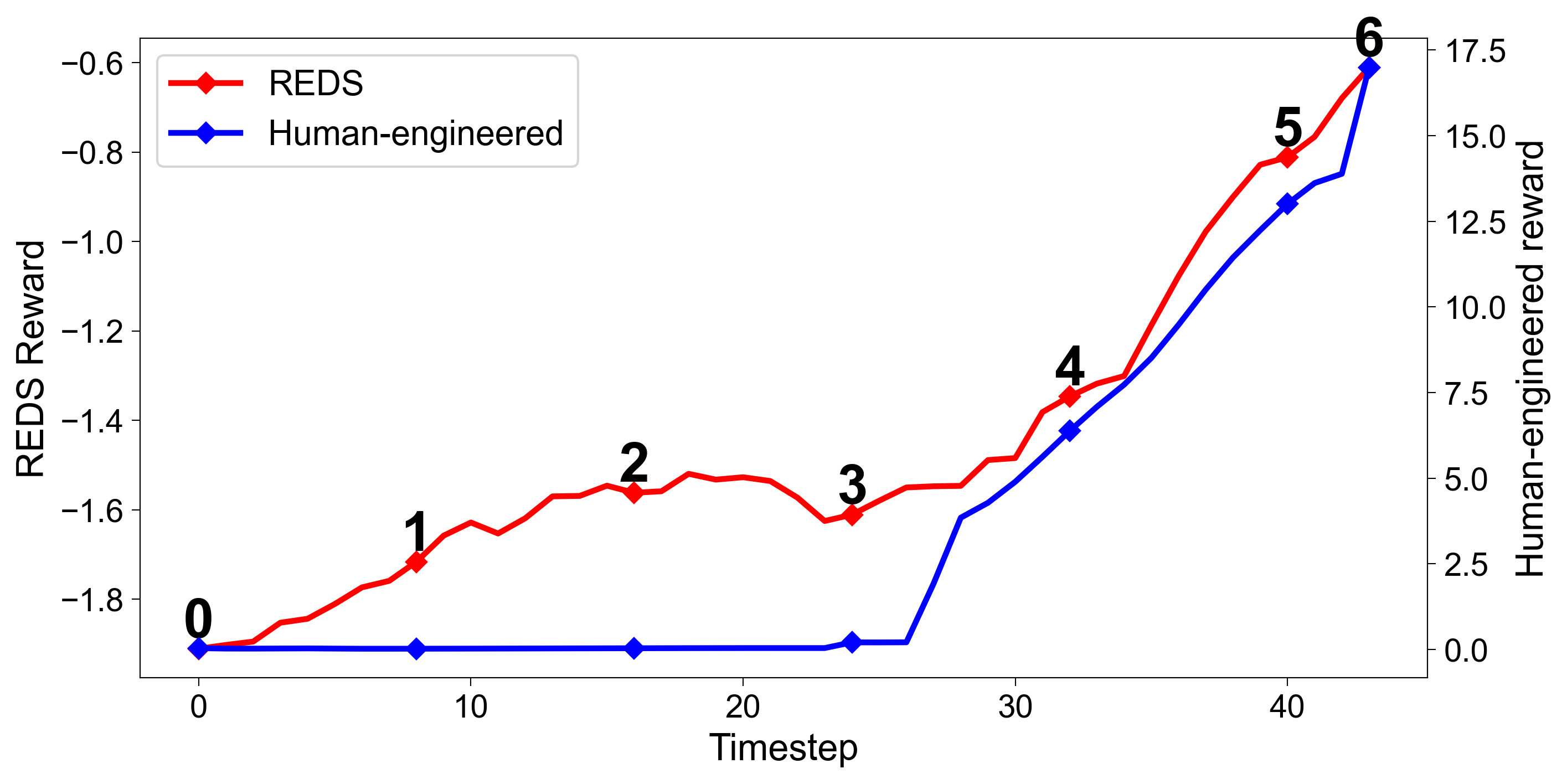}
         \label{fig:baseline_qual_peg_graph}
    \end{subfigure}
    \quad
    \begin{subfigure}[b]{.48\textwidth}
         \centering
         \includegraphics[width=\textwidth]{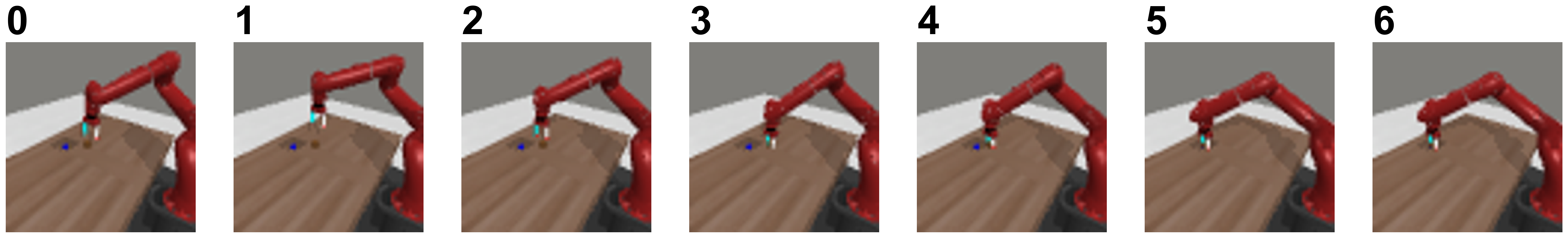}
         \includegraphics[width=\textwidth]{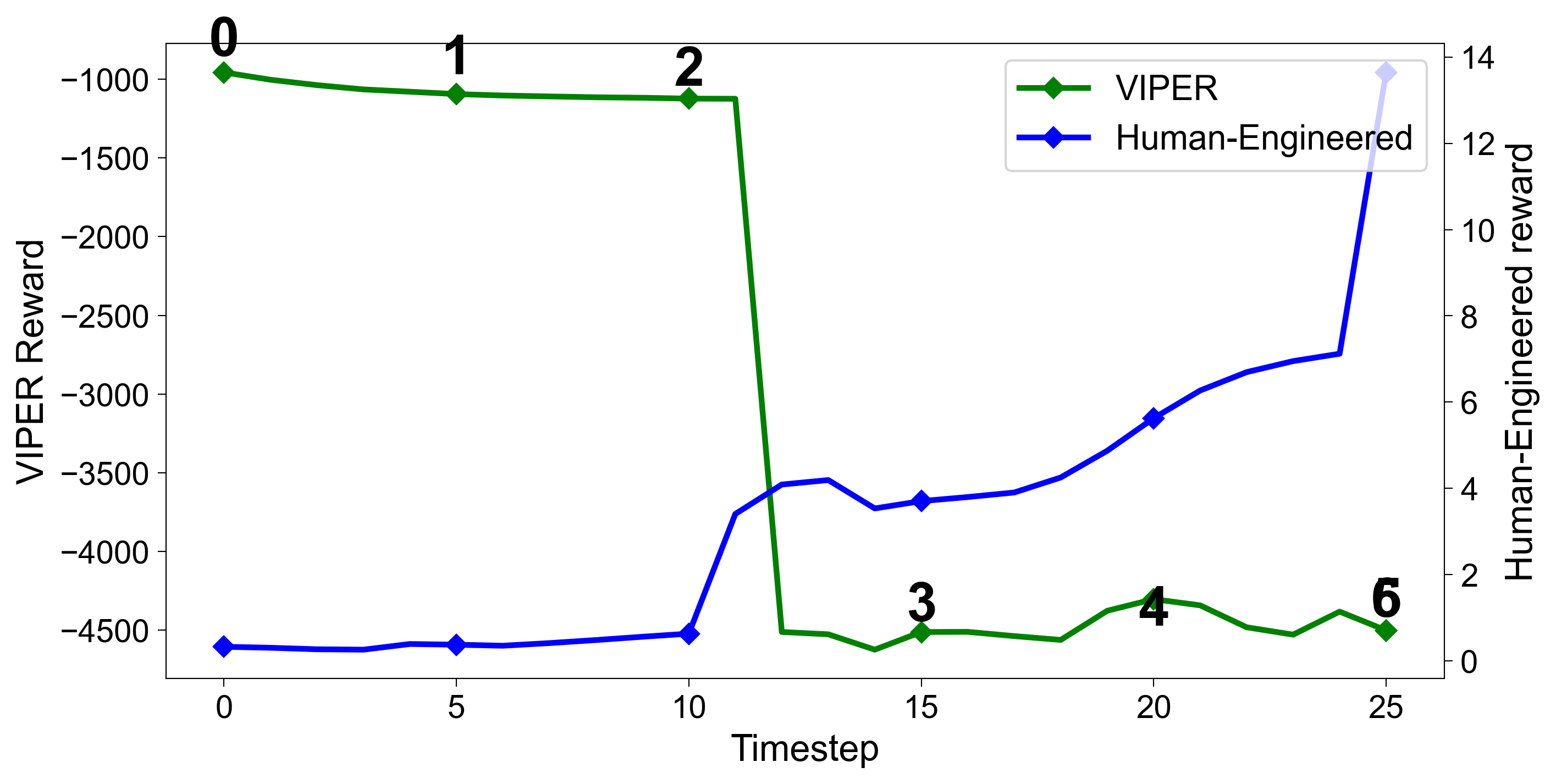}
         \includegraphics[width=\textwidth]{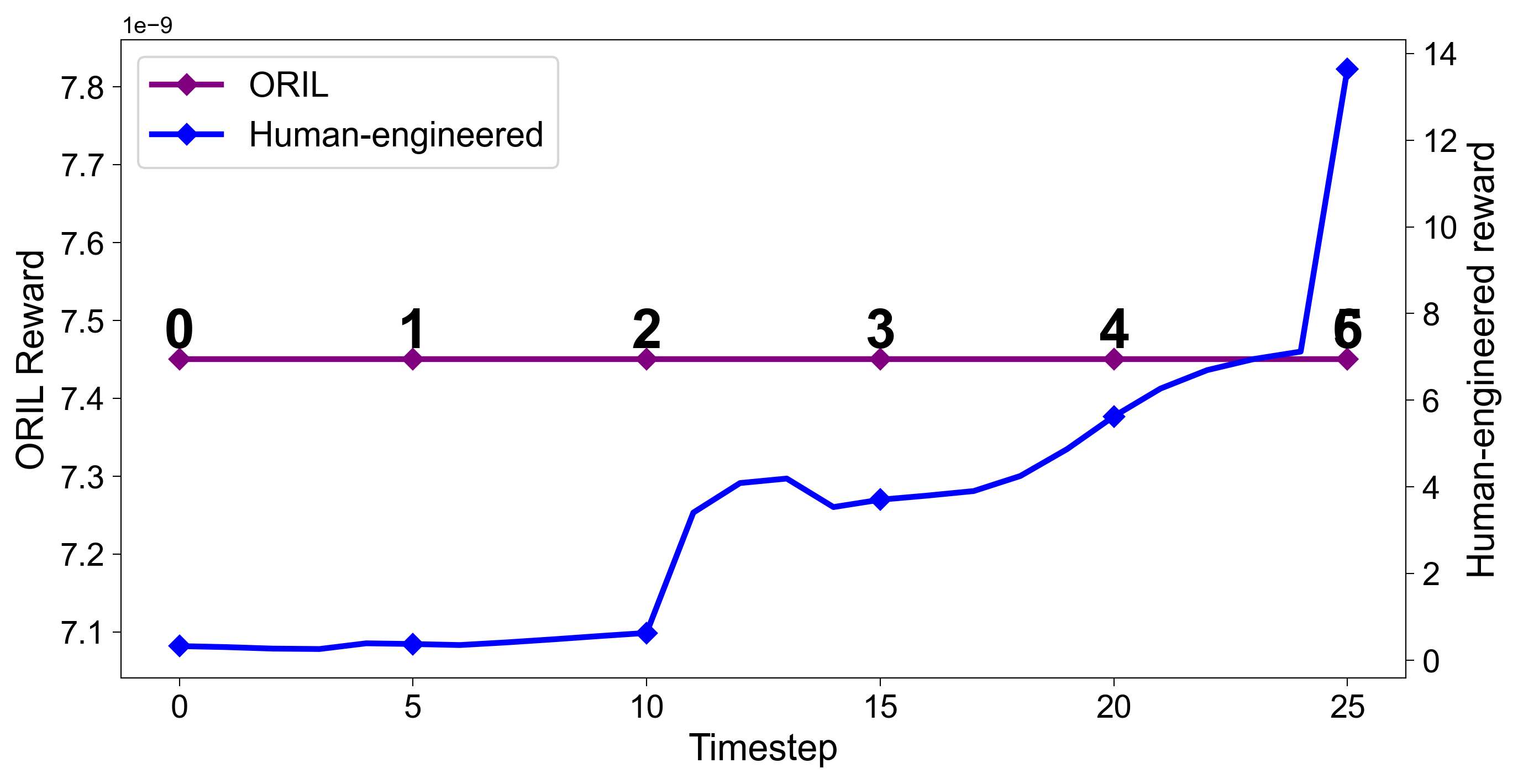}
         \includegraphics[width=\textwidth]{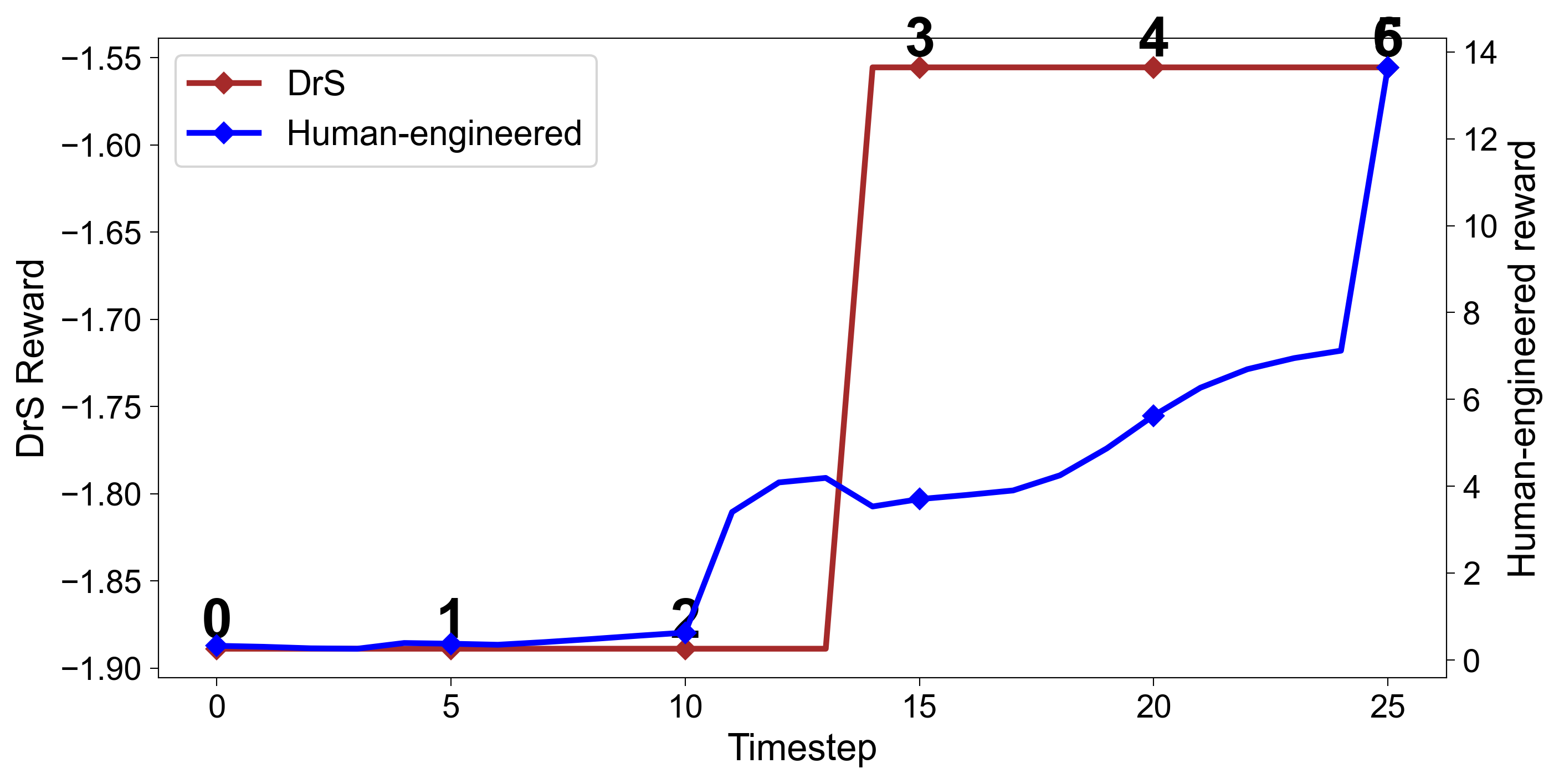}
         \includegraphics[width=\textwidth]{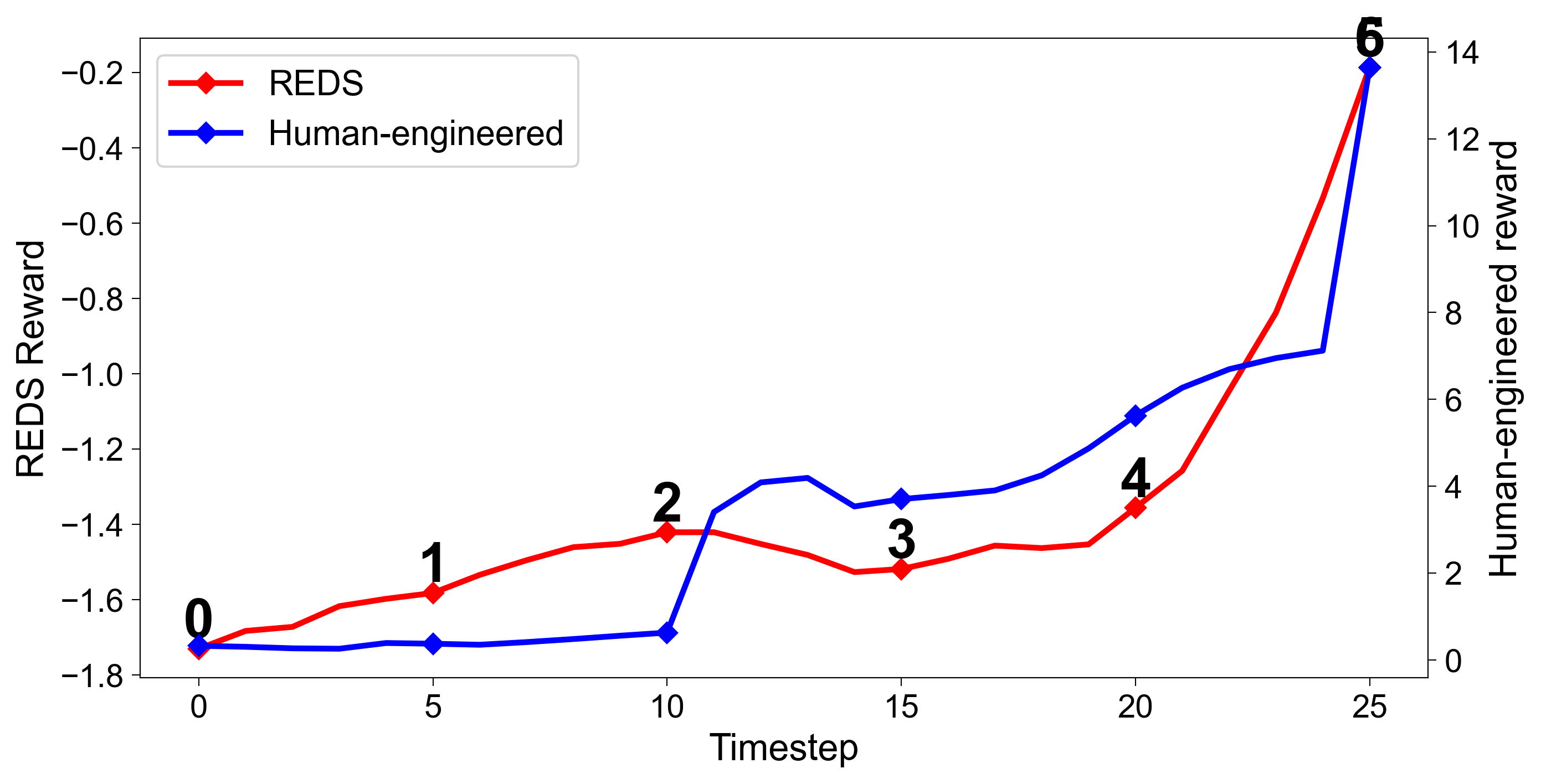}
         \label{fig:baseline_qual_sweep_graph}
    \end{subfigure}
    \caption{
    Qualitative results of VIPER~\citep{VIPER}, ORIL~\citep{oril}, DrS~\citep{drs}, and REDS (Ours) in Peg Insert Side (left), and Sweep Into (right) from Meta-world~\cite{yu2020meta}. We visualize several frames above the graph and mark them with a diamond symbol. 
    }
    \label{fig:qual_mw2}
\end{figure}
     \begin{figure} [h] \centering
    \begin{subfigure}[b]{0.95\textwidth}
         \centering
         \includegraphics[width=.85\textwidth]{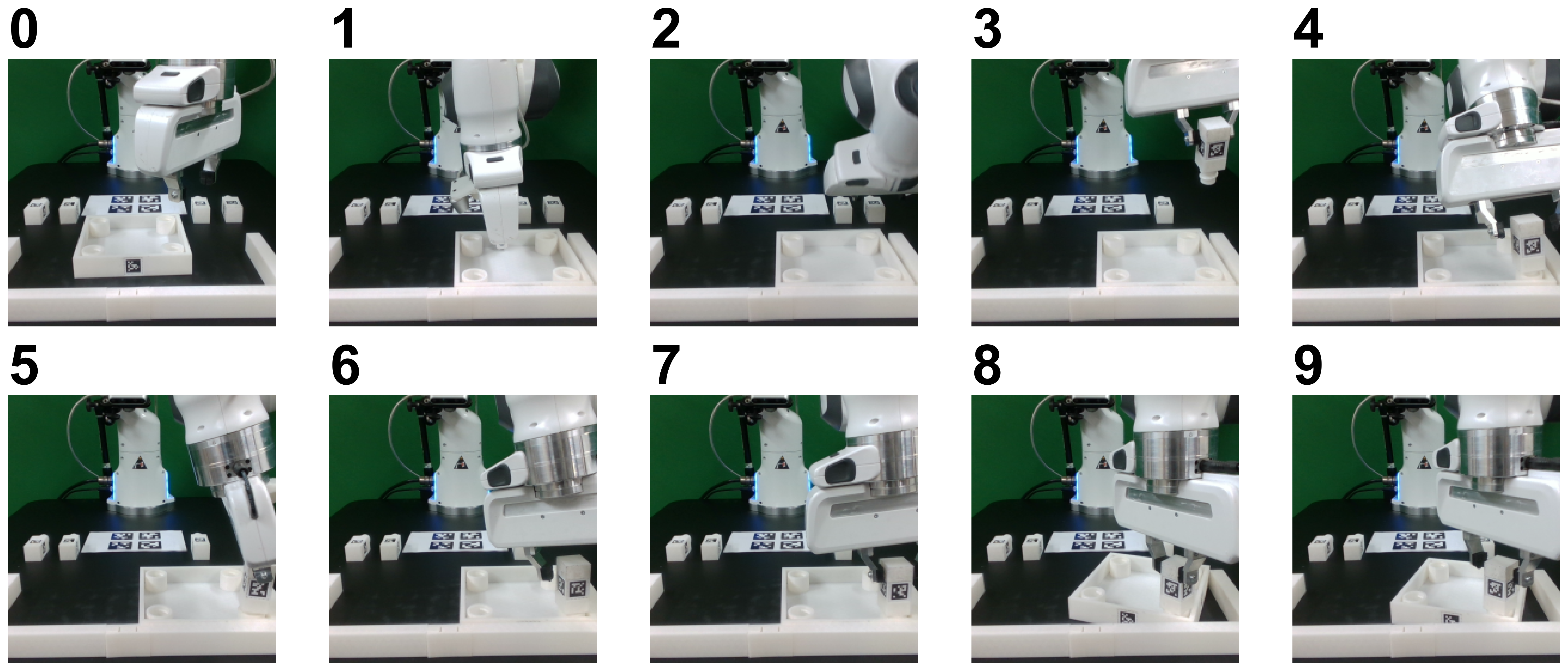}
         \includegraphics[width=\textwidth]{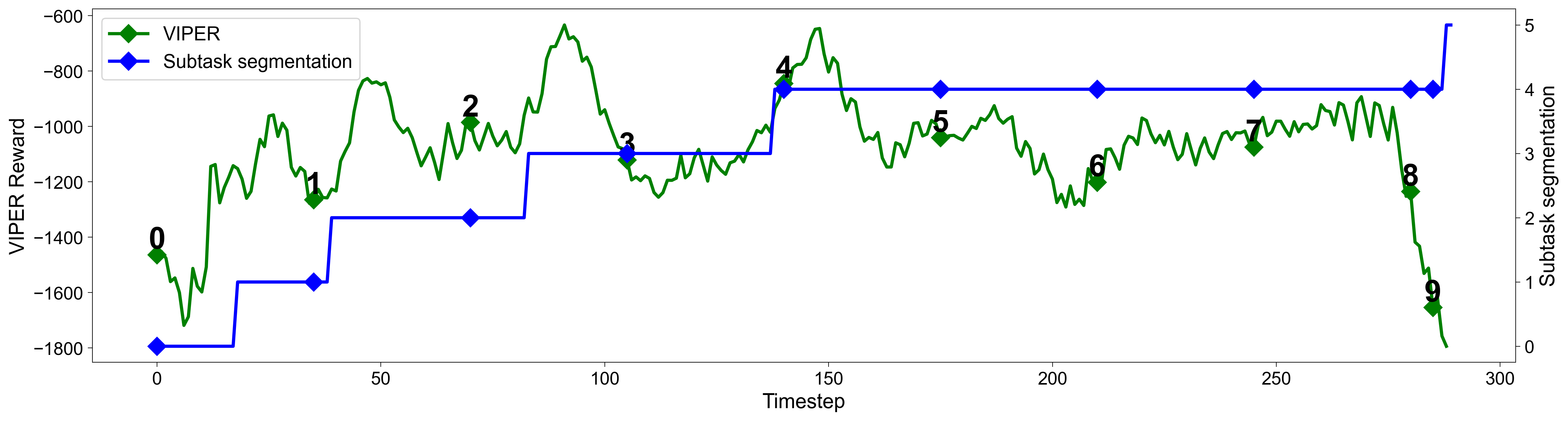}
         \includegraphics[width=\textwidth]{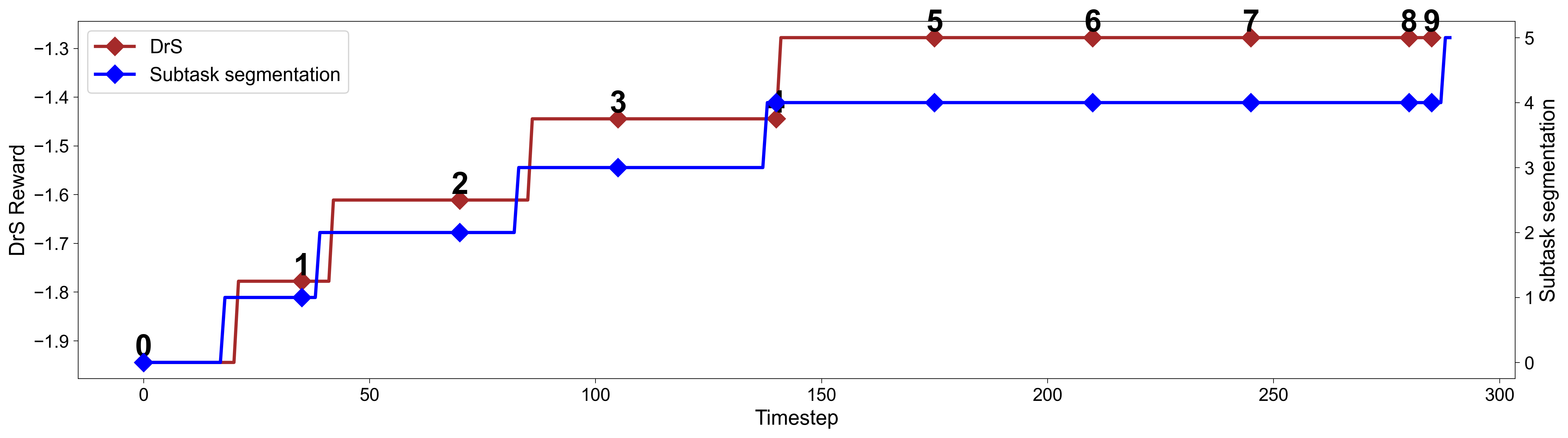}
         \includegraphics[width=\textwidth]{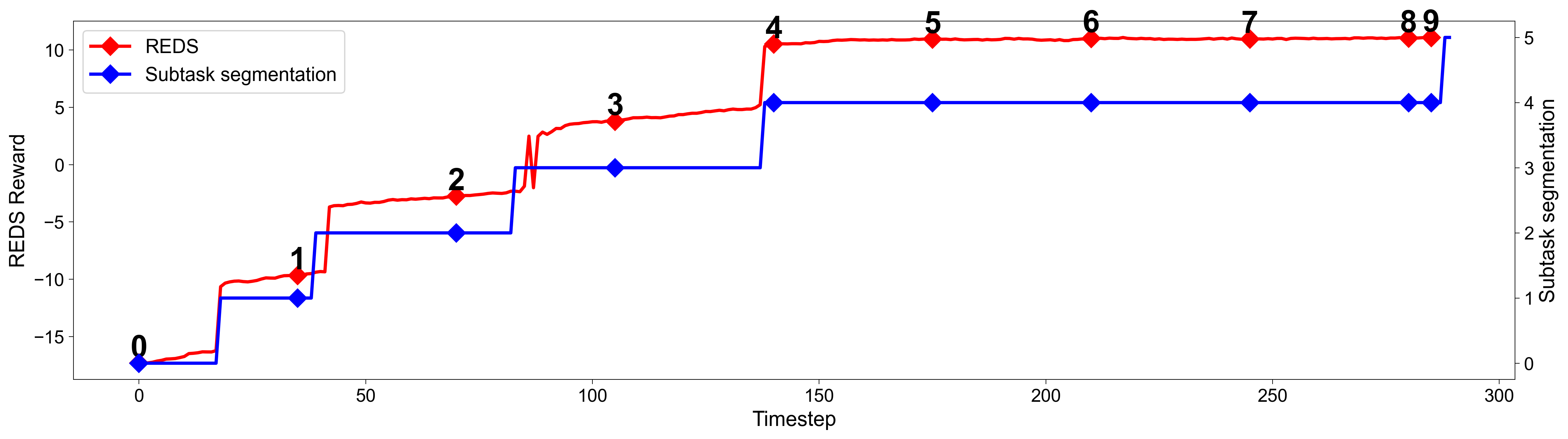}
         \label{fig:qual_fb_one_leg_appendix}
    \end{subfigure} \\
    \vspace{-0.1in}
    \begin{subfigure}[b]{0.95\textwidth}
        \centering
        \includegraphics[width=.19\textwidth]{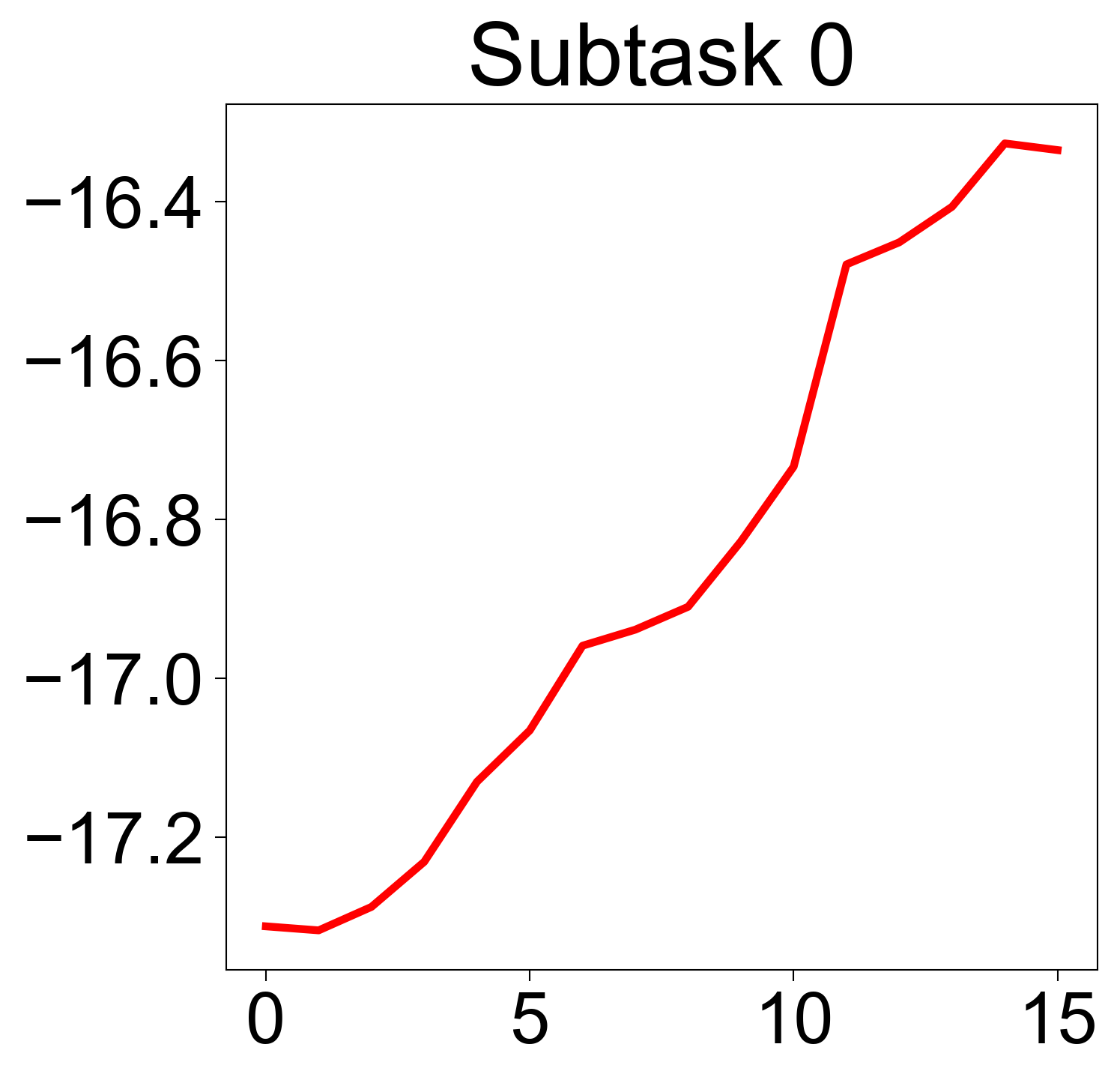}
        \includegraphics[width=.19\textwidth]{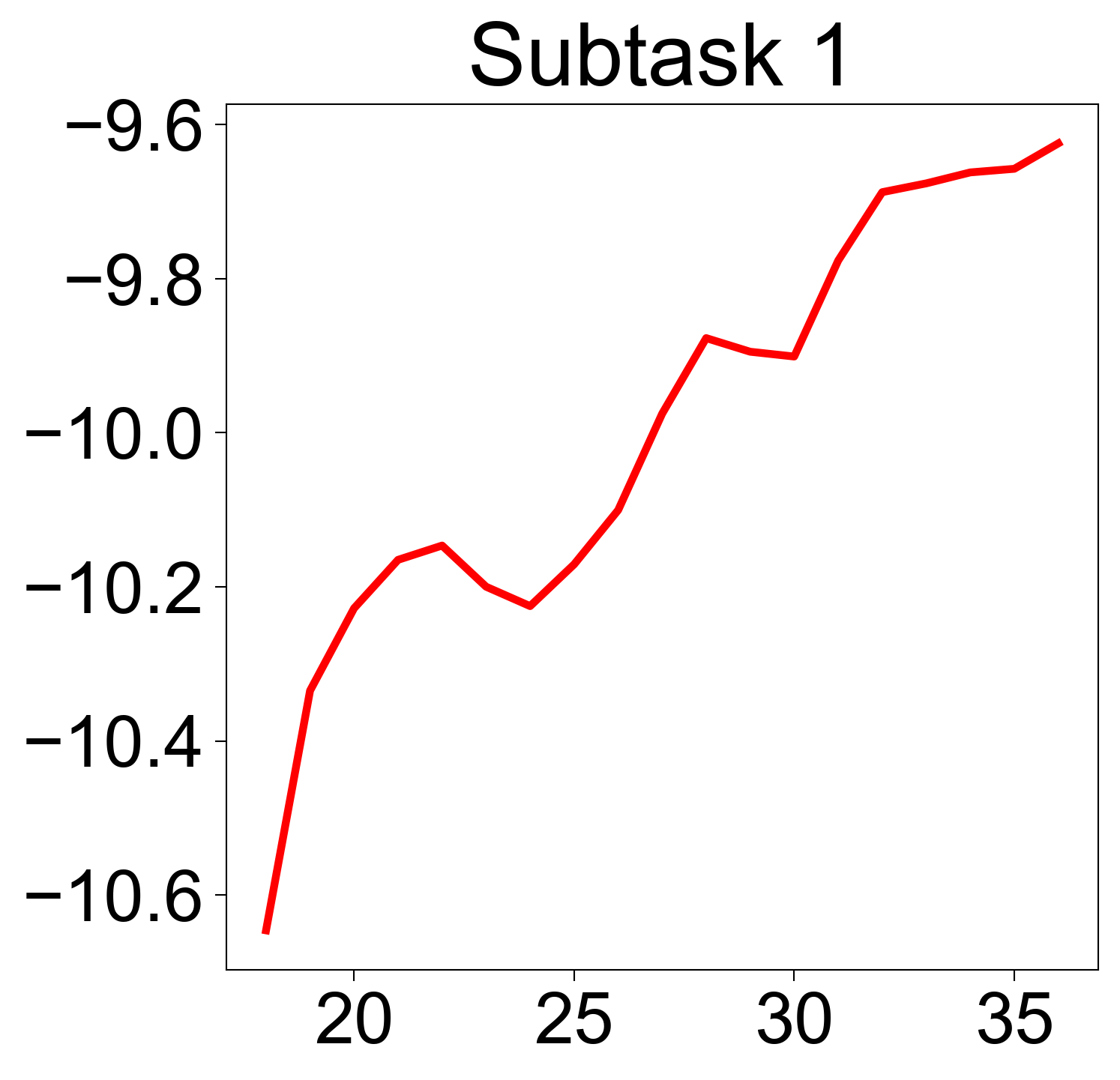}
        \includegraphics[width=.19\textwidth]{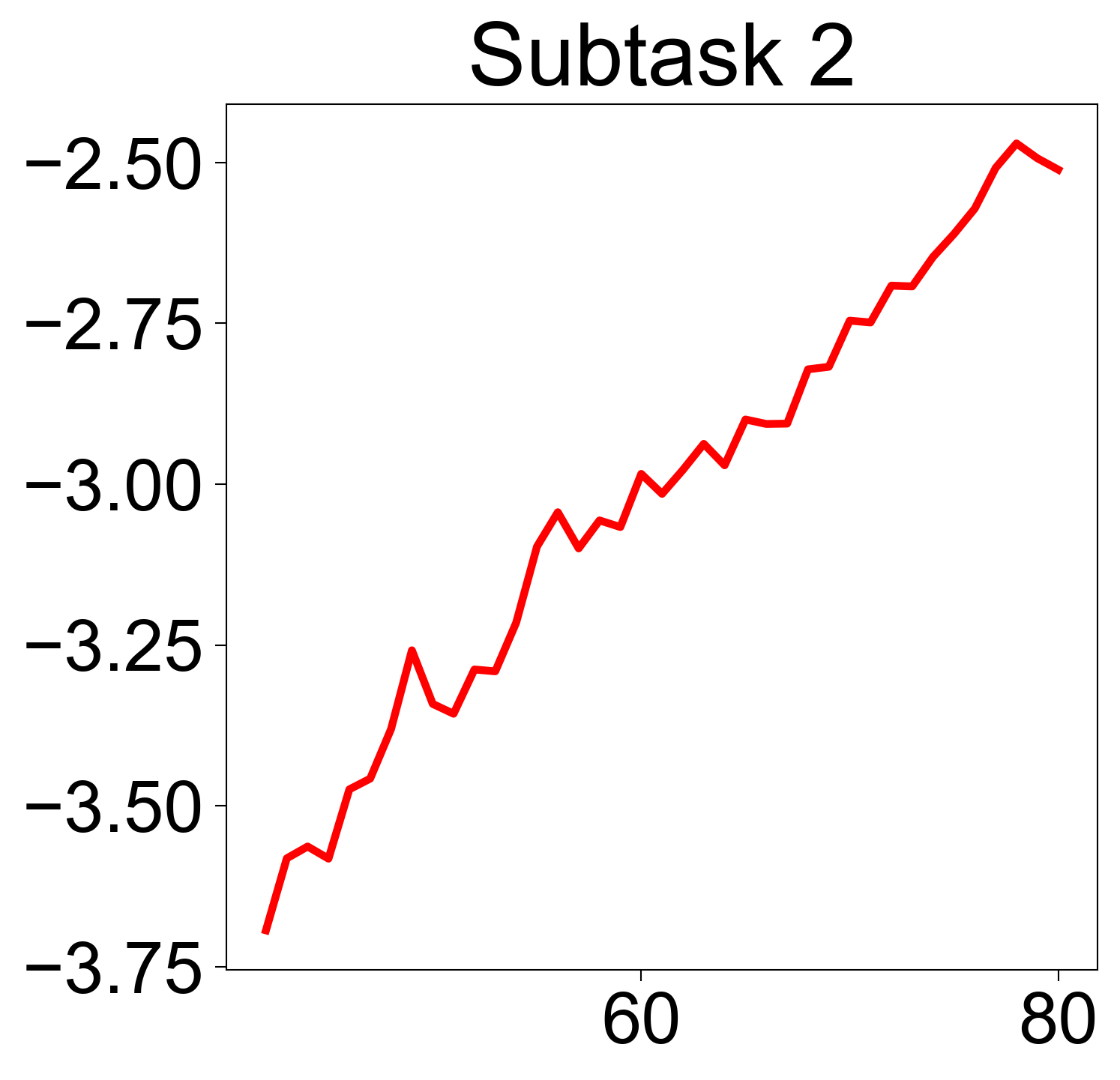}
        \includegraphics[width=.19\textwidth]{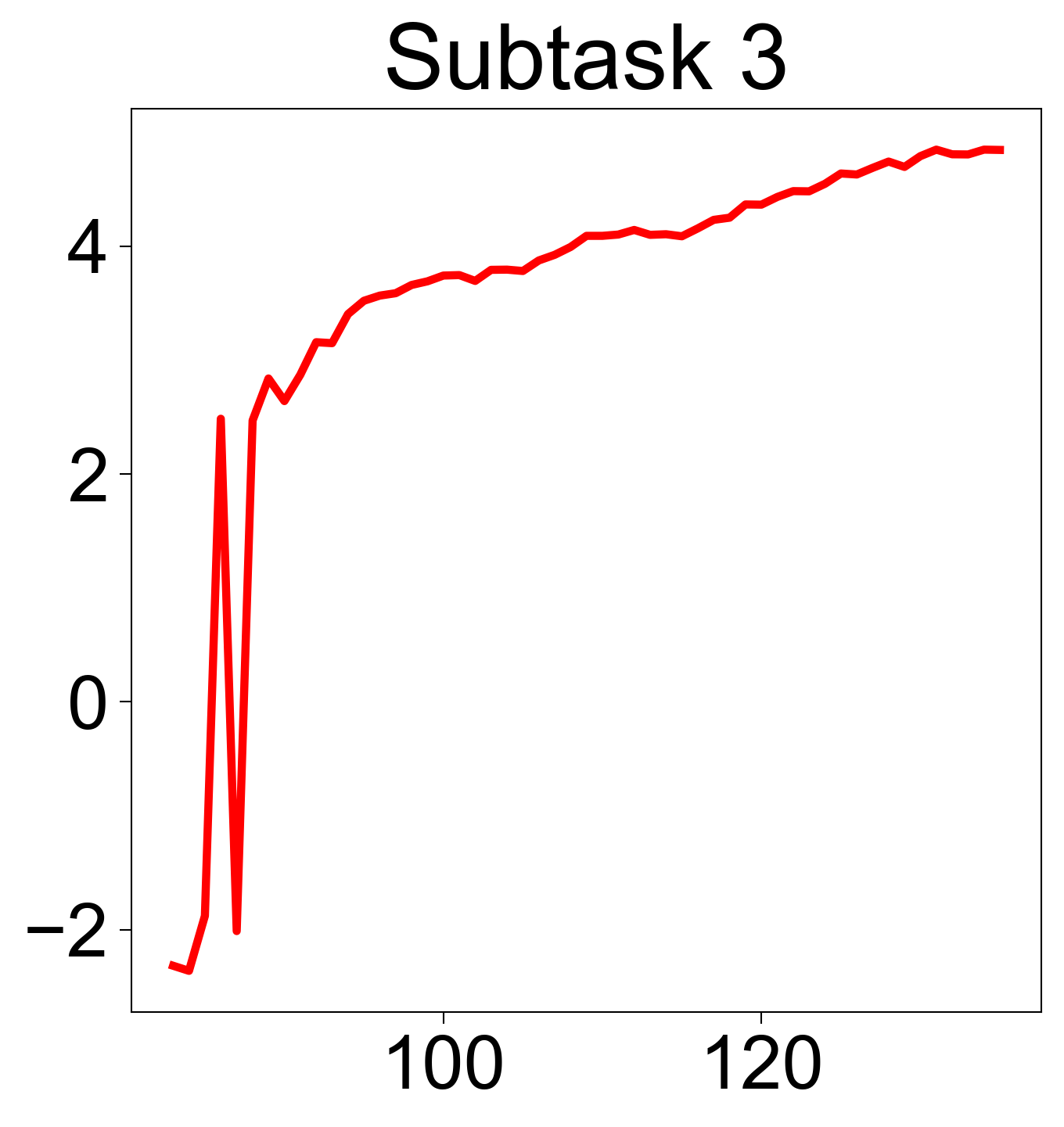}
        \includegraphics[width=.19\textwidth]{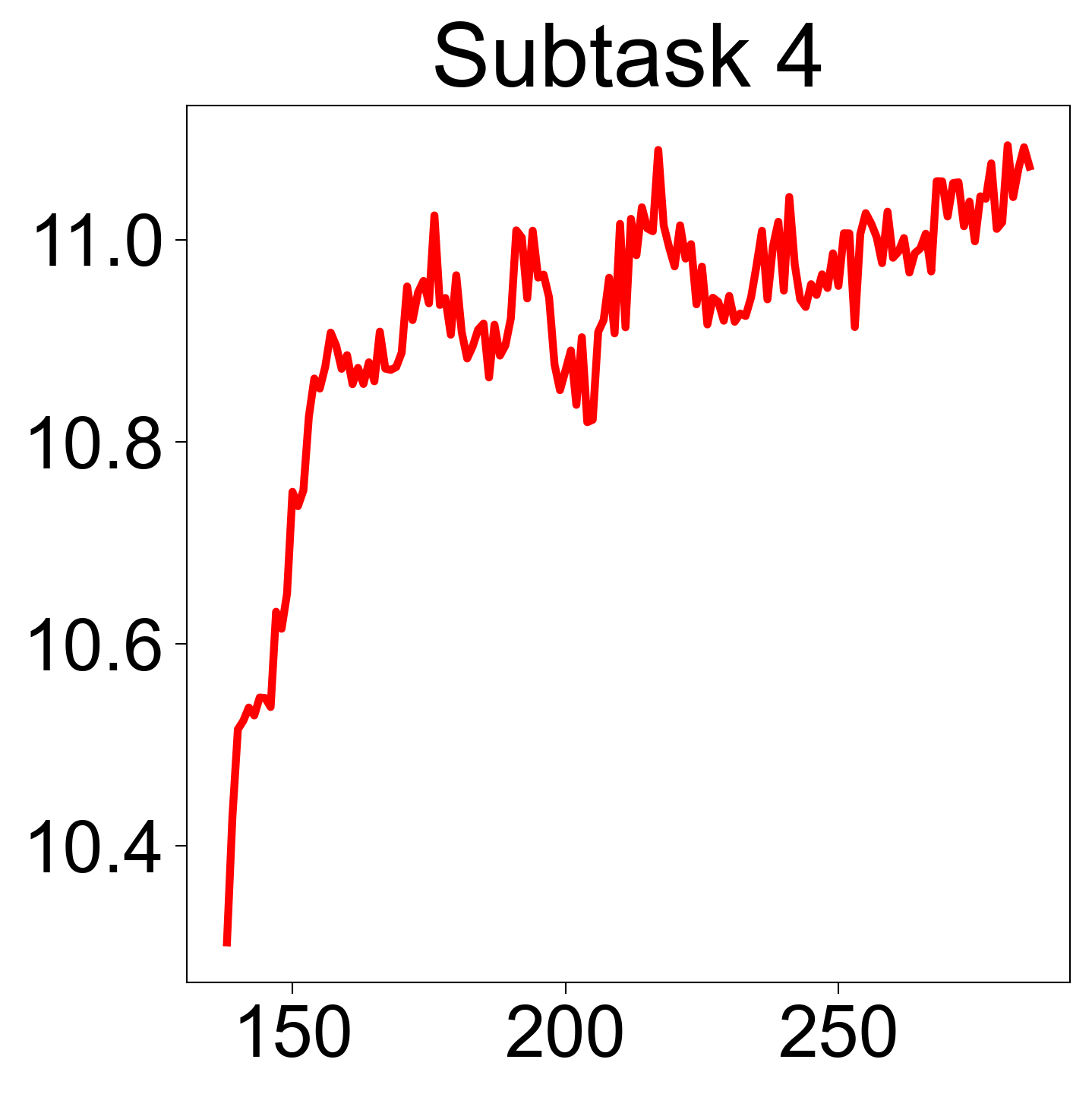}
    \end{subfigure}
    \vspace{-0.05in}
    \caption{
        Qualitative results of VIPER~\citep{VIPER}, DrS~\citep{drs}, and REDS (Ours) in One Leg from FurnitureBench~\citep{furniturebench}. We visualize several frames above the graph and mark them with a diamond symbol. VIPER, which does not utilize subtask information, assigns lower rewards to later subtasks, making agents stagnate in earlier phases. While DrS uses ground-truth subtask information from the environment, it produces sparse reward signals within each subtask. In contrast, REDS provides subtask-aware signals in subtask transitions and generates progressive reward signals (see the bottom figure zoomed in for each subtask).
    }
    \label{fig:baseline_qual_fb}
\end{figure}

\clearpage

\section{Additional Experiments}

\begin{figure}[t]
    \centering
    \begin{subfigure}[b]{0.32\textwidth}
        \centering
        \includegraphics[width=\textwidth]{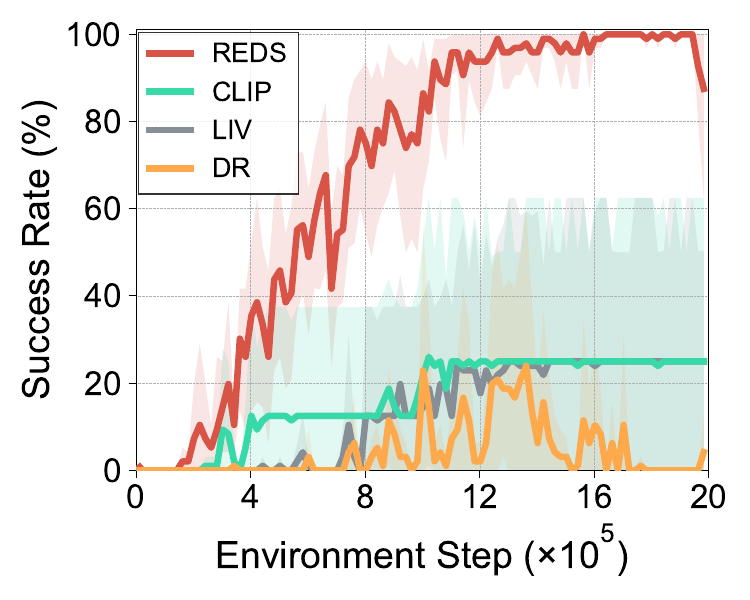}
        \vspace{-0.22in}
        \caption{Additional baselines}
        \label{fig:mw_abl_vlm_rewards}
    \end{subfigure}
    \begin{subfigure}[b]{0.32\textwidth}  
        \centering 
        \includegraphics[width=\textwidth]{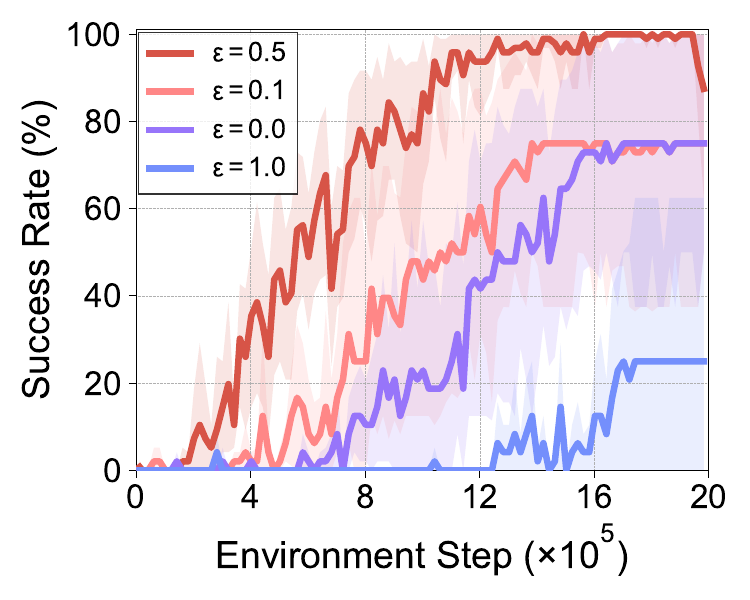}
        \vspace{-0.22in}
        \caption{hyperparameter $\epsilon$}
        \label{fig:mw_abl_epic_eps}
    \end{subfigure}
    \begin{subfigure}[b]{0.32\textwidth}   
        \centering 
        \includegraphics[width=\textwidth]{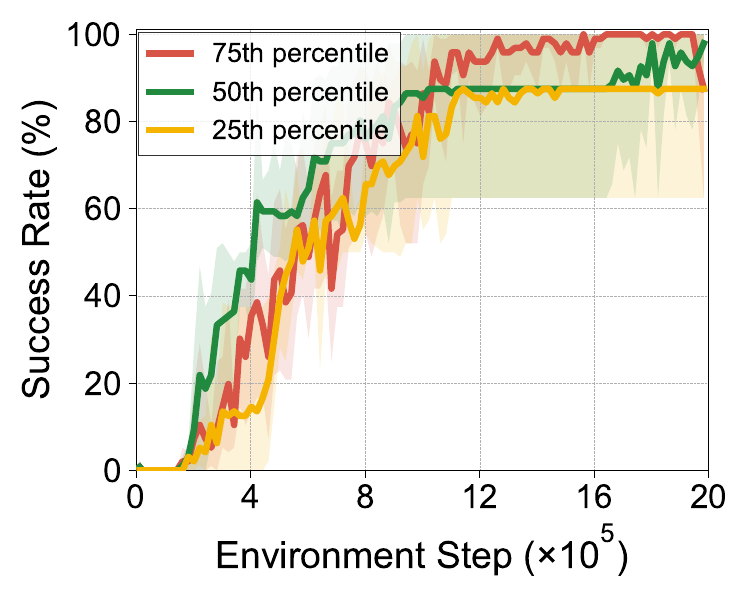}
        \vspace{-0.22in}
        \caption{Threshold $T_{U}$}
        \label{fig:mw_abl_fail_threshold}
    \end{subfigure}

    \caption{
    Learning curves for two Meta-world~\citep{yu2020meta} robotic manipulation tasks, measured by success rate (\%). The solid line and shaded regions show the mean and stratified bootstrap interval across 4 runs.}
    \vspace{-0.2in}
    \label{fig:additional_ablation}
\end{figure}
\paragraph{Comparison with additional baselines} We first compare REDS with additional baselines, CLIP~\citep{CLIP} and LIV~\citep{ma2023liv}, utilizing the distance between visual observation and text instructions for generating rewards. It's important to note that these models cannot infer subtasks in online interaction unlike REDS; therefore, we use other text instructions describing how to solve the whole task. (refer to Table~\ref{tbl:task_description_clip} for details). Figure~\ref{fig:mw_abl_vlm_rewards} shows that REDS significantly outperforms baselines, indicating that providing detailed signals aware of subtasks is crucial for better RL performance. Additionally, we compare \metabbr with Diffusion Reward (DR)~\citep{diffusion_reward}, which utilizes conditional entropy from a video diffusion model as a reward signal. Our findings indicate that REDS also significantly outperforms DR. This is attributed to the fact that DR does not explicitly incorporate subtask information, which is essential for generating context-aware rewards in long-horizon tasks. These results further emphasize the advantage of REDS in handling tasks requiring precise subtask guidance.

\paragraph{Effect of scaling progressive reward signals} In Figure~\ref{fig:mw_abl_epic_eps}, we examine the effect of $\epsilon$ scaling the regularization for progressive reward signals in Equation~\ref{eq:progressive}. We observe that $\epsilon = 0.5$ shows the best performance, while smaller values relatively weaken helpful progressive signals, and larger values degrade the reward function by reducing the accuracy in inferring subtasks.

\paragraph{Effect of threshold $T_{U}$ for subtask inference} We present experimental results using various threshold $T_{U}$ in Figure~\ref{fig:mw_abl_fail_threshold}. We observe that a lower percentile threshold exhibits lower RL performance. These results indicate that a lower percentile threshold allows more observations to be classified as successful; however, this can lead to misleading subtask identification, resulting in decreased RL performance.

\begin{wrapfigure}{r}{0.4\textwidth}
    \vspace{-0.15in}
    \captionof{table}{
        Precision in identifying subtasks of REDS on 50 unseen expert demonstrations and 50 unseen suboptimal demonstrations.
    }\label{tbl:subtask_identification}
    \vspace{-0.05in}
    \centering\resizebox{0.4\textwidth}{!}{
        \begin{tabular}{@{}cccc@{}}
        \toprule
        Fine-tuning & Expert  & Suboptimal & Total   \\ \midrule
            \xmark     & 94.49\% & 70.90\%  & 82.70\% \\
            \cmark     & 92.56\% & 91.49\% & 92.03\% \\ 
        \bottomrule
        \end{tabular}
    }
    \vspace{-0.1in}
\end{wrapfigure}
\paragraph{Subtask identification ability of REDS} To assess the subtask identification capability of REDS, we measure its precision before and after fine-tuning with additional suboptimal demonstrations. This evaluation involves using 50 unseen expert demonstrations and 50 suboptimal demonstrations sampled from the replay buffer of DreamerV3 agents that were trained with a human-engineered reward. Table~\ref{tbl:subtask_identification} shows that the precision is comparable for expert demonstrations for both agents; however, there is a significant increase in precision for suboptimal demonstrations after fine-tuning. This improvement in precision results in enhanced RL performance, as illustrated in Figure~\ref{fig:mw_abl_refinement}.

\clearpage
\begin{wrapfigure}{r}{0.5\textwidth}
    \captionof{table}{
        EPIC~\citep{EPIC} distance (lower is better) between learned reward functions and subtask segmentations in unseen data.
    }\label{tbl:epic_sparse}
    \vspace{-0.05in}
    \centering\resizebox{0.5\textwidth}{!}{
        \begin{tabular}{@{}lcccc@{}}
        \toprule
            Task               & VIPER  & R2R & ORIL   & \metabbr (Ours)            \\ 
        \midrule
        Meta-world Door Open       & 0.6017 & 0.5731 & 0.7017 & \textbf{0.4870}          \\
        Meta-world Push            & 0.6293 & 0.7014 & 0.7094 & \textbf{0.5129} \\
        Meta-world Peg Insert Side & 0.6384 & 0.6021 & 0.7001 & \textbf{0.4381}         \\
        Meta-world Sweep Into      & 0.6179 & 0.6584 & 0.7011 & \textbf{0.4293}                \\
        \bottomrule
        \end{tabular}
    }
    \vspace{-0.1in}
\end{wrapfigure}
\paragraph{Additional EPIC measurements} To further validate the efficacy of our method, we measure the EPIC distance between learned reward functions and subtask segmentations in Meta-world environments. Note that we report the result with the same set of unseen demonstrations used in Section~\ref{sec:alignment}. In Table~\ref{tbl:epic_sparse}, we observe that REDS exhibits significantly lower EPIC distance than baselines across all tasks, consistently supporting the claims from the experiments in the main text.

\end{document}